\documentclass[times]{TRR}

\usepackage{subcaption}
\usepackage{graphicx}

\usepackage{moreverb,url}

\usepackage[colorlinks,bookmarksopen,bookmarksnumbered,citecolor=red,urlcolor=red]{hyperref}

\newcommand\BibTeX{{\rmfamily B\kern-.05em \textsc{i\kern-.025em b}\kern-.08em
T\kern-.1667em\lower.7ex\hbox{E}\kern-.125emX}}

\def\volumeyear{2020}

\begin{document}

\runninghead{Pramanik et al.}

\title{Computational and experimental design of fast and versatile magnetic soft robotic low Re swimmers}

\author{R Pramanik\affilnum{1,6}, M Park\affilnum{2}, Z Ren\affilnum{2,3}, M Sitti\affilnum{2,4,5}, RWCP Verstappen\affilnum{1} and PR Onck\affilnum{6}}

\affiliation{\affilnum{1}Computational \& Numerical Mathematics Group, Bernoulli Institute for Mathematics, Computer Science \& Artificial Intelligence, University of Groningen, Netherlands\\
\affilnum{2}Physical Intelligence Department, Max Planck Institute for Intelligent Systems, Stuttgart, Germany
\\
\affilnum{3}School of Mechanical Engineering and Automation, Beihang University, Beijing, China
\\
\affilnum{4}Institute for Biomedical Engineering, ETH Zurich, Zurich, Switzerland
\\
\affilnum{5}School of Medicine and College of Engineering, Koc University, Istanbul, Turkey
\\
\affilnum{6}Micromechanics Group, Zernike Institute for Advanced Materials, University of Groningen, Netherlands}

\corrauth{PR Onck, p.r.onck@rug.nl}

\begin{abstract}
Miniaturized magnetic soft robots have shown extraordinary capabilities of contactless manipulation, complex path maneuvering, precise localization, and quick actuation, which have equipped them to cater to challenging biomedical applications such as targeted drug delivery, internal wound healing, and laparoscopic surgery. However, despite their successful fabrication by several different research groups, a thorough design strategy encompassing the optimized kinematic performance of the three fundamental biomimetic swimming modes at miniaturized length scales has not been reported till now. Here, we resolve this by designing magnetic soft robotic swimmers (MSRSs) from the class of helical and undulatory low Reynolds number (Re) swimmers using a fully coupled, experimentally calibrated computational fluid dynamics model. We study (and compare) their swimming performance, and report their steady-state swimming speed for different non-dimensional numbers that capture the competition by magnetic loading, non-linear elastic deformation and viscous solid-fluid coupling. We investigate their stability for different initial spatial orientations to ensure robustness during real-life applications. Our results show that the helical 'finger-shaped' swimmer is, by far, the fastest low Re swimmer in terms of body lengths per cycle, but that the undulatory 'carangiform' swimmer proved to be the most versatile, bi-directional swimmer with maximum stability.
\end{abstract}



\maketitle

\section{Introduction}
Small-scale stimuli-responsive soft materials are topics of intense scientific research due to their promising potential for challenging biomedical and microfluidic applications \cite{chen2018small,ouyang2022advances,hines2017soft}. A judicious mixing of external filler materials with elastomers yields smart flexible composites (also known as active elastica) that exhibit cross-domain energy transduction seamlessly and efficiently \cite{peng2021recent}. Therefore, these functional materials serve as promising candidates for state-of-the-art applications in soft robotics, stretchable electronics, biomedical engineering, and microfluidics \cite{gao2021synergizing,rich2018untethered,lin2023recent,yao2024multimodal,chen2023bioinspired}. In particular, soft robots find enormous use as devices for targeted drug delivery, cellular manipulation, and laparoscopy as cargo vehicles, grippers, endoscopes, capsules, etc. \cite{tripathi2024biomedical,ng2021locomotion,xia2022multicomponent}. Several researchers have used different actuation strategies to drive these active materials including magnetic \cite{li2023magnetic}, optical \cite{yin2021visible}, hydraulic \cite{joyee2019fully}, pneumatic \cite{zhang2019miniature}, shape memory \cite{huang2019highly}, and electrical fields \cite{shin2018electrically}.

Amongst these, the magnetic mode of actuation has been mostly adopted due to their intrinsic abilities of minimally invasive untethered (remote) actuation, high depth of penetration, and quick response time \cite{eshaghi2021design,xue2023magnetically}. Furthermore, this mode of actuation does not depend a lot upon the intermediate medium between the input source (external coils) and region of interest (hard-to-reach body sites) \cite{ebrahimi2021magnetic}. As a result, biochemical reactions are not triggered in the bio-fluids of the patient during surgical applications \cite{kim2022magnetic}. In addition, not only does a magnetic field offer a spatio-temporal control with high precision, but also provides a versatile untethered non-invasive approach to remote control robotic systems \cite{xia2024magnetic}. During their intended applications, these soft robots are often desired to operate within and swim through narrow viscous fluidic channels or confinements, thereby giving rise to complex large deformation fluid-structure interactions \cite{ren2021soft}. However, swimming at such microscopic length scales is difficult due to high viscous drag and negligible inertial effects \cite{abbott2009should}. Inspired by nature, researchers have identified a wide range of swimmers that have evolved over millions of years for efficient propulsion through similar fluidic environments \cite{fish2014evolution,higham2015turbulence}. At these low Reynolds number flows (Re $\ll$ 1), inertial forces do not (significantly) contribute to net propulsion \cite{lauga2009hydrodynamics,purcell1977life,garstecki2009swimming}. Therefore, natural lifeforms typically adopt diverse swimming strategies to propel themselves through viscid flows \cite{velho2021bank}. They employ different types of non-reciprocal motion to achieve spatial asymmetry \cite{lauga2011life}.

In nature, there are broadly three distinct swimming modes - helical, undulatory, and ciliary \cite{pramanik2024nature}. The undulatory swimmers either generate traveling body waves to push the surrounding fluid backward to move forward or employ an oar-like motion to sway the surrounding fluid for net propulsion. The helical swimmers typically develop a body chirality (twisting) and use a cork-screw motion to propel forward. And, the ciliary swimmers adopt a non-reciprocal motion consisting of asymmetric contraction (effective) and relaxation (recovery) phases to manipulate the surrounding fluid for swimming. Aquatic lifeforms such as bacteria, fish, midge larvae, etc. have optimized their swimming behavior over large evolutionary timescales. Microorganisms such as bacteria, spermatozoa, and ciliates use compliant hair-like tubular appendages (e.g., cilia and flagella) to manipulate the surrounding fluid for effective propulsion and other physiological processes \cite{guasto2012fluid,cicconofri2019modelling,pak2015theoretical}. E.g., spermatozoa travel through a mucus-laden path to reach the ovum for reproduction \cite{foo2008biofluid}. \textit{E. Coli}, single-flagellated \textit{Pseudomonas aeruginosa}, and peritrichously-flagellated \textit{Ensifer meliloti} use a helical cork-screw motion to swim through viscous fluids \cite{chattopadhyay2006swimming,vater2014swimming,gotz1987rhizobium}. Undulatory swimmers such as midge larvae generate flexural body waves to periodically displace the surrounding fluid for net propulsion \cite{schoeller2018flagellar,ishimoto2018human}. Ciliary swimmers such as \textit{Chlamydomonas reinhardtii}, \textit{Paramecium caudatum}, and jellyfish \textit{Ephyra} employ asymmetric contraction (effective) and relaxation (recovery) strokes to propel themselves \cite{chen2017swimming,schnitzler2022reversible,feitl2009functional}.

Owing to their efficient swimming strategies, they have often motivated researchers trying to develop soft robotic swimmers \cite{park2016phototactic,el2020optimal,magdanz2013development,cohen2010swimming,muralidharan2023bio}. In this regard, magnetically actuated miniaturized swimmers have been successfully fabricated and experimentally characterized to exhibit remote manipulation, precise path maneuvering, and quick actuation \cite{asghar2018mechanical,magdanz2020ironsperm,bhattacharjee2022bacteria,xu2021discrete}. Magnetic thin films have been subjected to oscillatory magnetic fields, resulting in undulatory swimmers that either generated traveling body waves or exhibited an oar-like motion to push the surrounding fluid for net propulsion \cite{manamanchaiyaporn2020magnetic,huang2019visual,zhang2018untethered,xu2019millimeter,zhang2015millimeter}. Chiral swimmers have displayed helical propulsion under rotating magnetic fields for effective swimming \cite{liu2017swimming,garstecki2009propulsion}. \textit{Ephyra} and \textit{Paramecium}-inspired magnetic robots have been developed that use a ciliary mode of propulsion to sweep a net surrounding fluid volume for thrust generation. A non-reciprocal magnetic field has been used to actuate the \textit{Paramecium}-inspired magneto-responsive swimmer, wherein the protruding cilia-shaped tubular appendages exhibited unequal contraction and relaxation strokes to propel forward \cite{kim2016fabrication}. An \textit{Ephyra}-inspired robotic swimmer with partially-magnetized lappets deformed distinctly within one swimming cycle under non-reciprocal magnetic fields to achieve fluidic propulsion \cite{ren2019magnetically}. Furthermore, switching between undulatory and helical swimming modes has been reported for a sperm-templated soft robot with a rigid magnetic head and a flexible passive tail \cite{khalil2018controllable}. Adaptive multi-modal locomotion with improved swimming performance has been reported for soft-bodied magneto-responsive swimmers that were able to generate undulatory body waves as well as develop chirality for effective propulsion through confined (narrow) fluid spaces \cite{ren2019multi}.

Although noteworthy developments have been made to successfully fabricate magnetic soft robotic swimmers (MSRSs) and demonstrate their advanced swimming modalities \cite{zhang2023small,xu2024deformation,gurbuz2023elastohydrodynamic,ren2023design,ren2024undulatory}, there is still a gap in the identification of the optimal propulsion strategy at low Re flows and the design of these MSRSs based on forward swimming speeds, precise path maneuverability, bi-directional swimming, and stability. This paper addresses these lacuna and provides a conclusive answer to these open questions by combining experiments and simulations.

For the experiments we use a manufacturing technique based on shape morphing and programmed magnetization \cite{ren2021soft,dong2020bioinspired,ren2023design} and for the modelling a fully coupled solid-fluid computational fluid dynamics model \cite{khaderi2012fluid,namdeo2014numerical,pramanik2023magnetic,zhang2022metachronal} by simultaneously solving for the magnetic forces, the non-linear solid deformation and the fluid flow. As the main readout, we use the normalized steady-state swimming speed in body lengths per cycle, blpc, and study this as a function of the non-dimensional magnetic (M$_\text{n}$) and fluid numbers (F$_\text{n}$). Furthermore, we investigate the influence of different aspect ratios and normalized magnetic lengths on their swimming behavior. The bi-directional swimming and steering modalities are discussed, while their stability under different spatial configurations is analyzed. Additionally, we investigate the flow field around these swimmers during one complete swimming cycle to understand its influence on swimming performance. Finally, we provide guidelines for designing a magnetic soft robotic swimmer that is most efficient in terms of its swimming speed, on-the-fly bi-directionality, maneuverability, and stability for critical biomedical applications. Our results identify helical finger-shaped swimmers to be the fastest and undulatory carangiform swimmers to be the most versatile for practical applications.

\begin{figure*}[htbp!]
\centering
\includegraphics[scale=0.33]{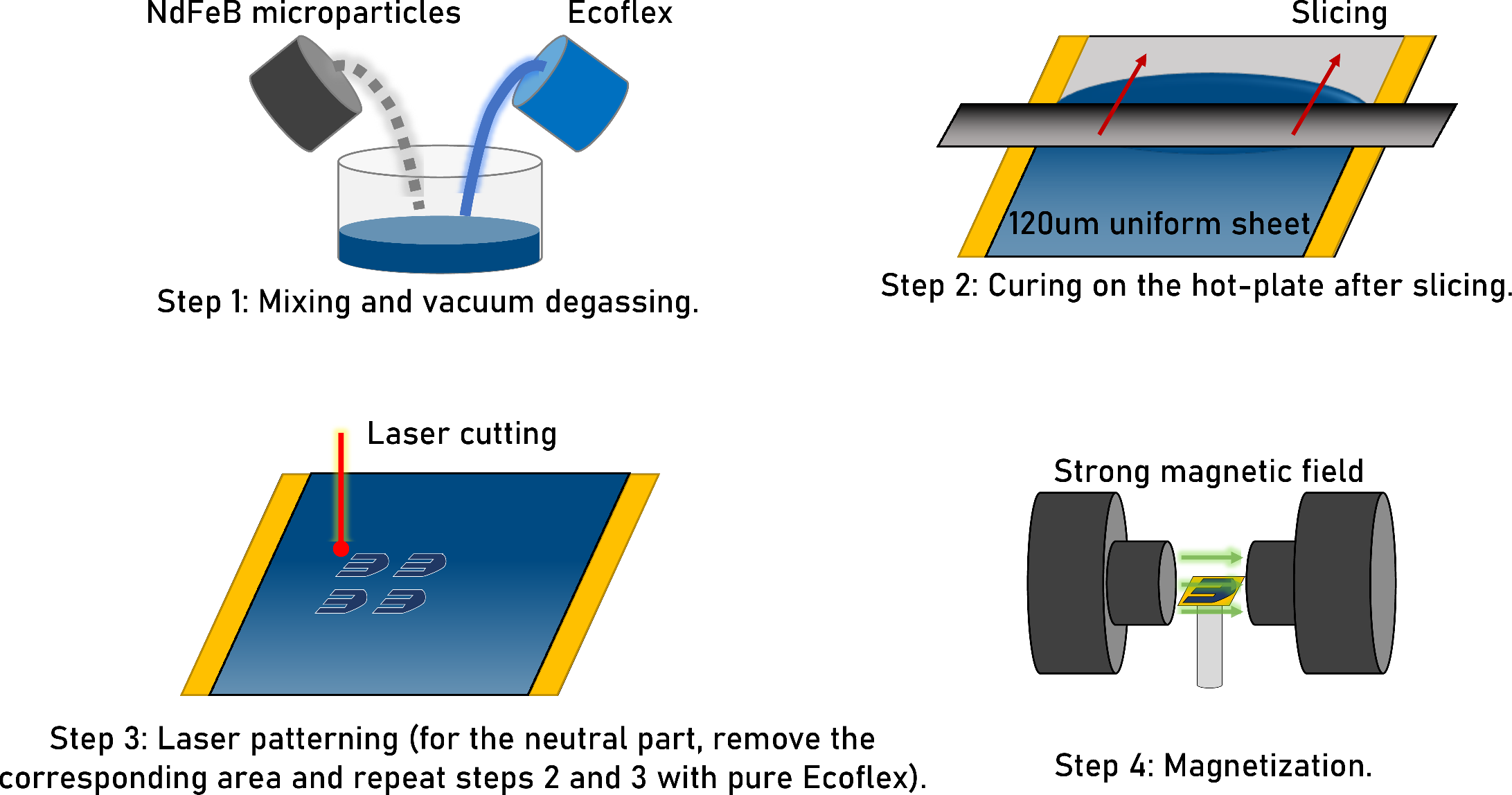}
\caption{Schematic representation of the fabrication procedure adopted for finger-shaped MSRS.}
\label{fig:fabrication}
\end{figure*}

\section{Results and Discussion}
\subsection{Fabrication of MSRSs}
The magnetic swimmers are fabricated using a casting method followed by laser patterning (see Fig. \ref{fig:fabrication}). Briefly, a mix of pre-cured silicone elastomer and NdFeB microparticles is prepared. This mixture is poured into molds defining the targeted film thickness and cured at 65$^{\circ}$C for 1h. Further, the cured magnetic soft materials are subjected to high magnetic fields (around 1T) to induce a magnetization profile \cite{lum2016shape}. Therefore, when these materials are (henceforth) subjected to relatively low magnetic fields (around 10mT), they shape-morph by deforming in a pre-defined manner. When these magnetic soft composite thin films are placed within a fluid medium, they displace the surrounding fluid to generate thrust for forward propulsion.

\begin{figure*}[htbp!]
\centering
\includegraphics[scale=0.4]{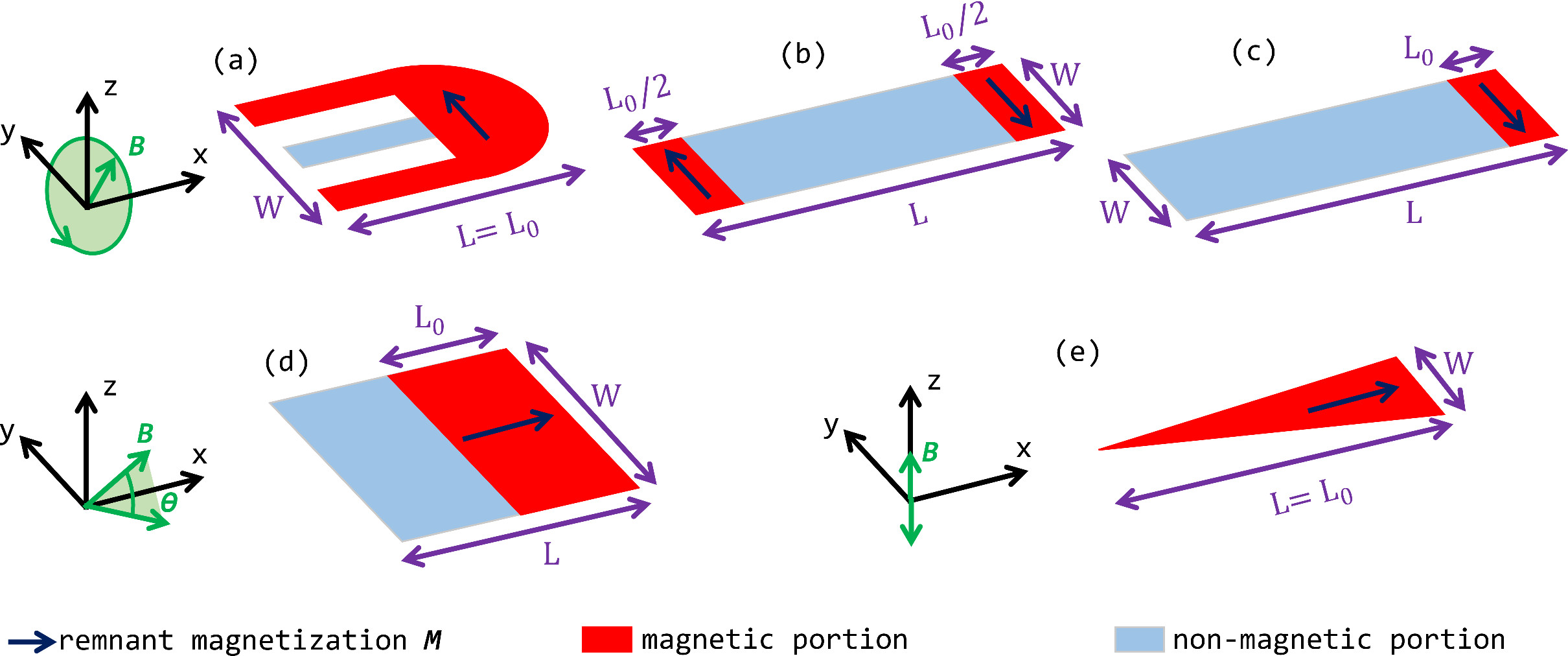}
\caption{Schematic representation of the magnetic soft robotic swimmers: (a) finger-shaped, (b) field-induced, (c) drag-induced, (d) carangiform-like, and (e) anguilliform-like. All the swimmers propel along the +ve x direction. The magnetic field vector B is represented by a green arrow, using a rotating magnetic field in (a) - (c) and an oscillating field in (d) and (e).}
\label{fig:swimmers}
\end{figure*}

\subsection{Experimental tests and computational modeling}
We fabricate the MSRSs and impart them with different profiles of remnant magnetization, such that they deform in a pre-defined manner subject to external magnetic fields (see Fig. \ref{fig:swimmers}). Although we have previously reported the computational design of a magnetically actuated jellyfish-inspired ciliary swimmer \cite{pramanik2023magnetic}, it was very challenging to fabricate the same for validation and experimental demonstration purpose. This was mainly because the swimmer exhibited a non-reciprocal contraction/relaxation stroke (typical of any ciliary swimmer) that had stringent requirements in terms of in-homogeneous (spatially non-uniform) remnant magnetization profiles and precise values of magnetic field (that had further dependencies on the fluid viscosity) for effective steady-state swimming and a net propulsion. Such accurate field settings/control and non-uniform magnetization profiles are extremely challenging to realize even with state-of-the-art laboratory facilities. Therefore, we propose a set of five MSRSs, out of which two are undulatory, and the rest are helical.

\begin{figure}[h]
\centering
\includegraphics[scale=0.33]{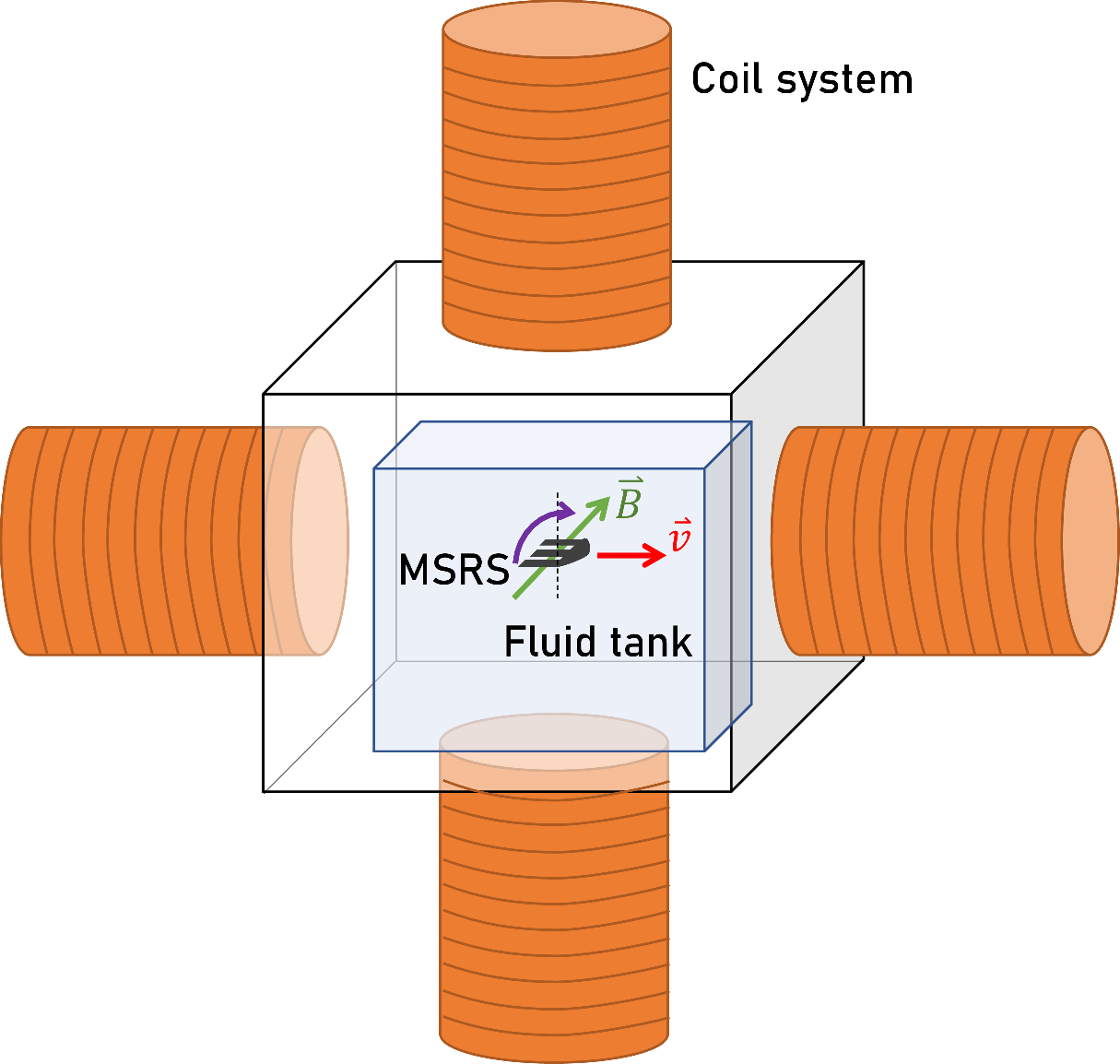}
\caption{Schematic representation of the experimental setup: the finger-shaped MSRS immersed within the fluid medium is subjected to magnetic fields for net propulsion.}
\label{fig:setup}
\end{figure}

Although there are variations in their geometry and magnetization profiles, they have a quite similar processing route. For this study, we fabricate one helical (finger-shaped) and one undulatory (carangiform-like) swimmer to compare with model predictions. A schematic representation of the magnetic field setup is shown in Fig. \ref{fig:setup}. To ensure a rational comparison of their swimming performance, the kinematics of these MSRSs are studied in terms of their normalized steady-state swimming speed (blpc) for variations in the non-dimensional magnetic (M$_\text{n}$) and fluid numbers (F$_\text{n}$), defined as M$_\text{n}$=$12{B}{M}\Bar{L}L_0/{Eh^2}$ and F$_\text{n}$=$12{\mu}\Bar{L}^3f_m/{Eh^3}$, with $E$ the Young's modulus, $\mu$ the viscosity of the fluid, \textit{B} the magnitude of the external magnetic field, \textit{M} the magnitude of the magnetization, $\Bar{L}$ the characteristic length, $L_0$ the magnetic segment length, $h$ the thickness, and $f_m$ the actuation frequency of \textit{B}. Here, we define blpc as the ratio of the steady-state swimming speed (\textit{c}) and the product of characteristic body length ($\Bar{L}$) and frequency ($f_\text{m}$): blpc=$c/\Bar{L}f_\text{m}$. Note that $\Bar{L}$ is computed as: $\Bar{L}=\sqrt{LW}$, where, \textit{L} and \textit{W} are the length and width of the swimmer, respectively.

\begin{figure*}[htbp!]
\centering
\includegraphics[scale=0.4]{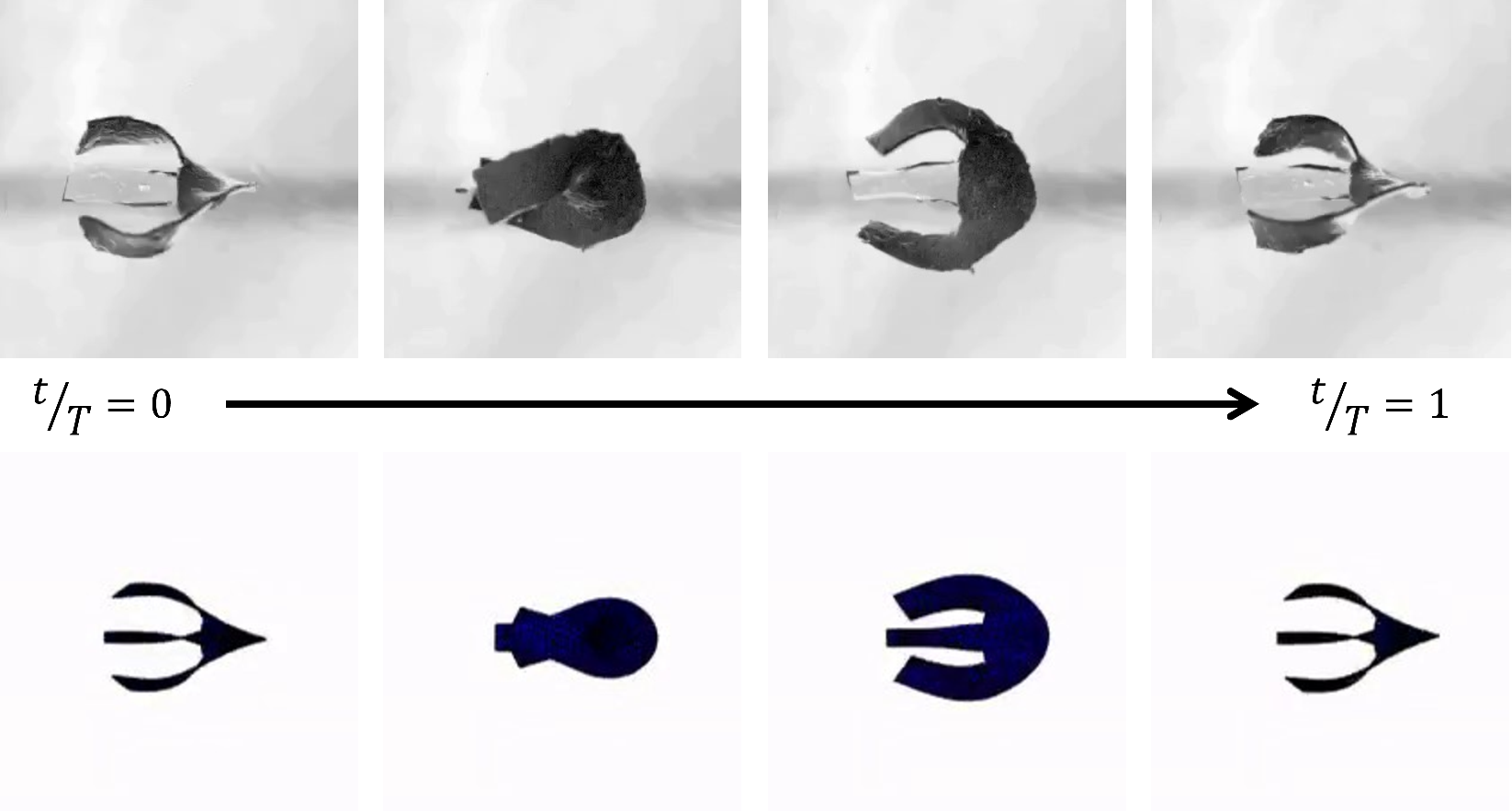}
\caption{Comparison of the chronological snapshots of the experimental observations (top row) with the model predictions (bottom row) for the finger-shaped helical swimmer during one swimming cycle. Here, \textit{t} and \textit{T} represent current time instant and cycle time period, respectively. For movies of the swimmers, see the Supplementary Information.}
\label{fig:snapshotsfinger}
\end{figure*}

\begin{figure*}[h]
\captionsetup[subfigure]{justification=centering}
     \centering
    \begin{subfigure}{\columnwidth}
         \centering
         \includegraphics[scale=0.45]{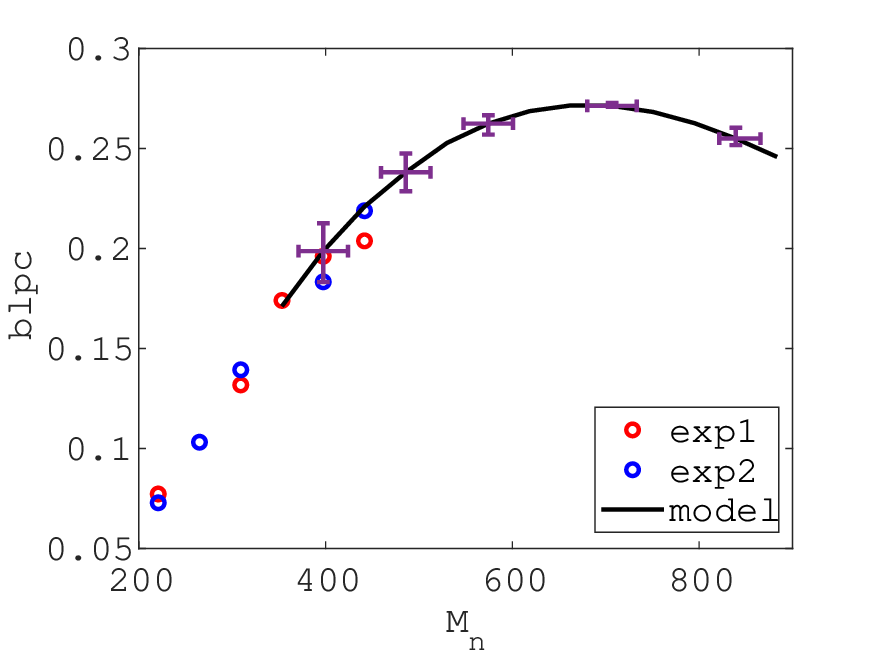}
         \caption{F$_\text{n}$=20}
         \label{fig:fingerexpmodelfn1}
     \end{subfigure}
     \hfill
    \begin{subfigure}{\columnwidth}
         \centering
         \includegraphics[scale=0.45]{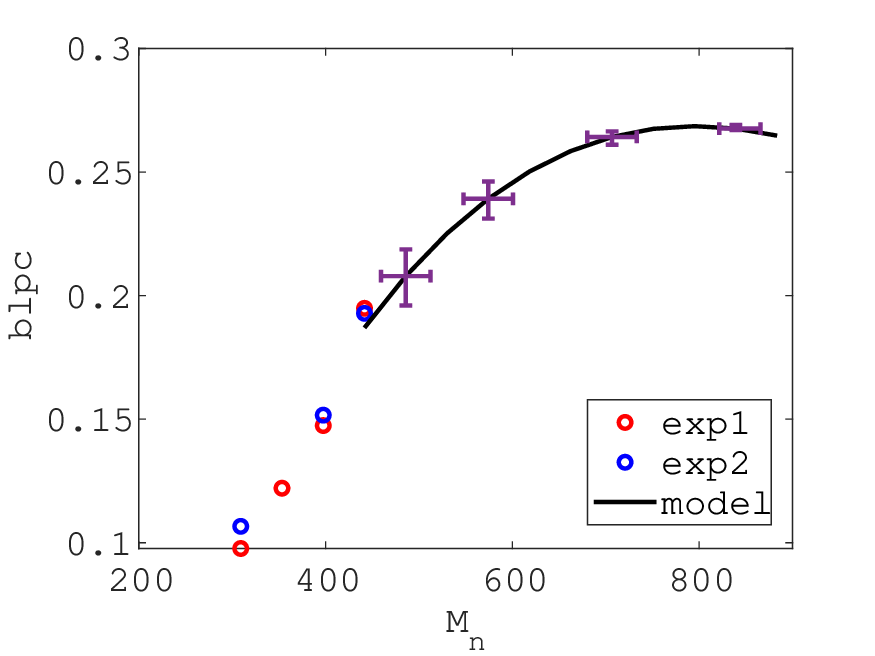}
         \caption{F$_\text{n}$=30}
         \label{fig:fingerexpmodelfn2}
     \end{subfigure}
        \caption{Comparing experiments and model: swimming speed vs. M$_\text{n}$ for (a) F$_\text{n}$=20 and (b) F$_\text{n}$=30 for the finger-shaped helical MSRS. The red and blue circles denote two separate experiments, the computational data points are represented by two error bars denoting the standard deviation. The numerical data is fitted by a regression curve (indicated by a black solid line).}
        \label{fig:expmodelfinger}
\end{figure*}

The finger-shaped (helical) swimmer has a uniform distribution of the external magnetic filler materials (i.e., a uniform remnant magnetization profile). This MSRS is subjected to rotating uniform magnetic fields \textit{B} that have the axis of rotation same as the swimmer body axis. Consequently, magnetic torques \textit{N} are generated, and defined as $N = M \times B$. These external body torques are further imposed upon the solid mechanics model of the swimmer. The (finger-like) outer flaps deform like cantilevers, while the central (inner) flap does not undergo remarkable deformation (and only curls slightly); we presume that the latter simply provides additional stability to the swimmer. The flaps push the surrounding fluid backwards rendering itself a net propulsion along the forward direction (see Fig. \ref{fig:snapshotsfinger}). The magnitude of the magnetic field is varied to perform a parametric sweep for M$_\text{n}$. For different values of fluid viscosity (and therefore F$_\text{n}$), we observe that the experimentally measured swimming performances (blpc) are in agreement with the model estimates (see Fig. \ref{fig:expmodelfinger}).

\begin{figure*}[htbp!]
\centering
\includegraphics[scale=0.4]{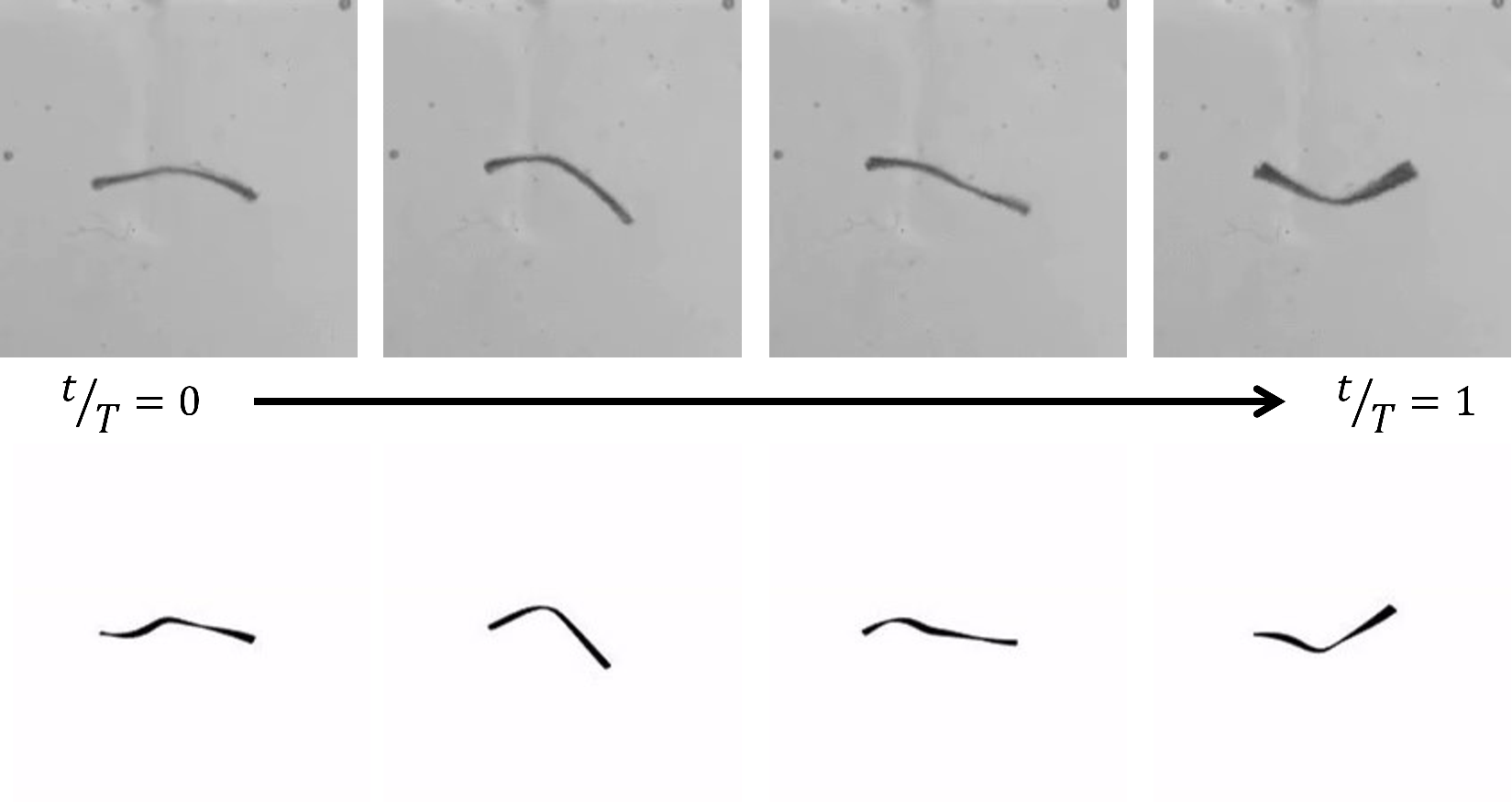}
\caption{Comparison of the chronological snapshots of the experimental observations (top row) with the model predictions (bottom row) for the carangiform-like undulatory swimmer during one swimming cycle. Here, \textit{t} and \textit{T} represent current time instant and cycle time period, respectively. For movies of the swimmers, see the Supplementary Information.}
\label{fig:snapshotscarangiform}
\end{figure*}

\begin{figure*}[h]
\captionsetup[subfigure]{justification=centering}
     \centering
     \begin{subfigure}[b]{0.49\textwidth}
         \centering
         \includegraphics[scale=0.45]{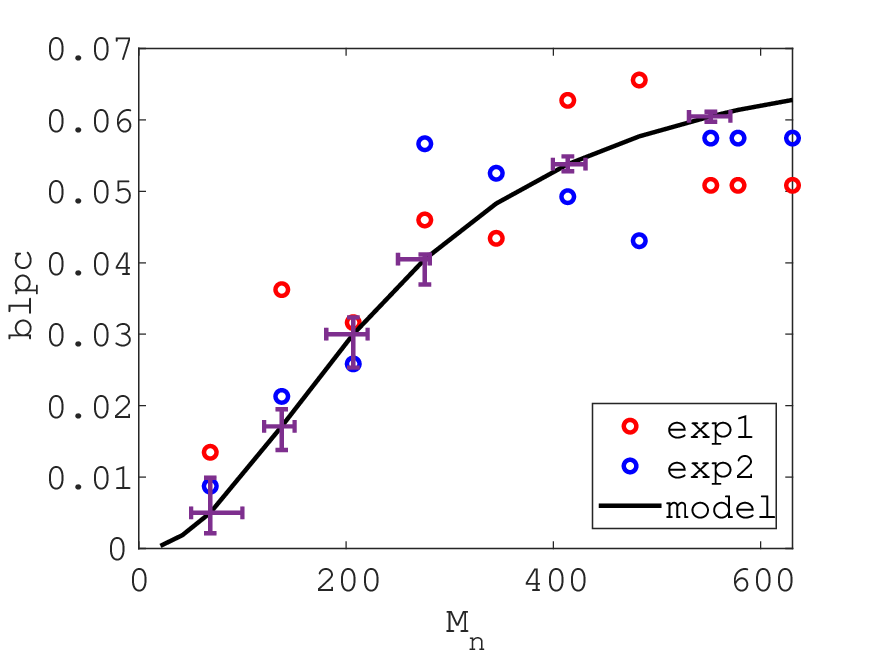}
         \caption{F$_\text{n}$=60}
         \label{fig:carangiformexpmodelfn1}
     \end{subfigure}
     \hfill
     \begin{subfigure}[b]{0.49\textwidth}
         \centering
         \includegraphics[scale=0.45]{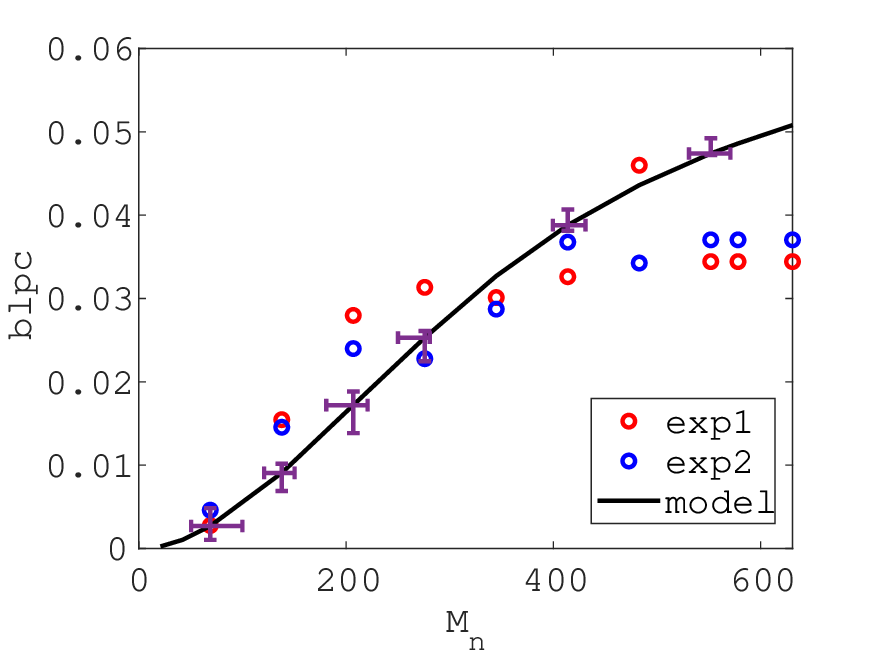}
         \caption{F$_\text{n}$=90}
         \label{fig:carangiformexpmodelfn2}
     \end{subfigure}
        \caption{Comparing experiments and model: swimming speed vs. M$_\text{n}$ for (a) F$_\text{n}$=60 and (b) F$_\text{n}$=90 for the carangiform-like undulatory MSRS. The red and blue circles denote two separate experiments, the computational data points are represented by two error bars denoting the standard deviation. The numerical data is fitted by a regression curve (indicated by a black solid line).}
        \label{fig:expmodelcarangiform}
\end{figure*}

The carangiform-like (undulatory) swimmer that swims using an oar motion is fabricated. A segment of the MSRS is uniformly magnetized (active portion), while the rest of the swimmer body has no magnetic properties (i.e., passive). This partially magnetized elastica (PME) is subjected to uniform directed oscillatory magnetic fields. The swimmer's active portion follows the magnetic field almost synchronously; however, the passive portion is not actuated. Rather, under structural elasticity and as a natural consequence of FSIs, it lags in phase compared to the active portion \cite{namdeo2013swimming}. This results in an oar-like (non-reciprocal) motion (see Fig. \ref{fig:snapshotscarangiform}). It is because of this phase lag, that the spatial symmetry is broken, resulting in a net propulsion. In the case of this magnetic swimmer as well,  we observe that the experimentally measured swimming performances (blpc) are in agreement with the model predictions (see Fig. \ref{fig:expmodelcarangiform}).

\subsection{Parametric study}
We compare the kinematic performance of individual MSRSs based on their blpc for variation in the other non-dimensional numbers such as M$_\text{n}$, F$_\text{n}$, normalized magnetic length $L_\text{0}/L$, and aspect ratio ($L/W$). We do this to estimate the highest swimming speeds for different system parameters including geometry, stiffness, viscosity, magnetic length, remnant magnetization, magnetic field, and actuation frequency - all of which are captured within the non-dimensional numbers \cite{pramanik2023magnetic}.

\begin{figure*}
\captionsetup[subfigure]{justification=centering}
     \centering
     \begin{subfigure}[b]{0.49\textwidth}
         \centering
         \includegraphics[scale=0.45]{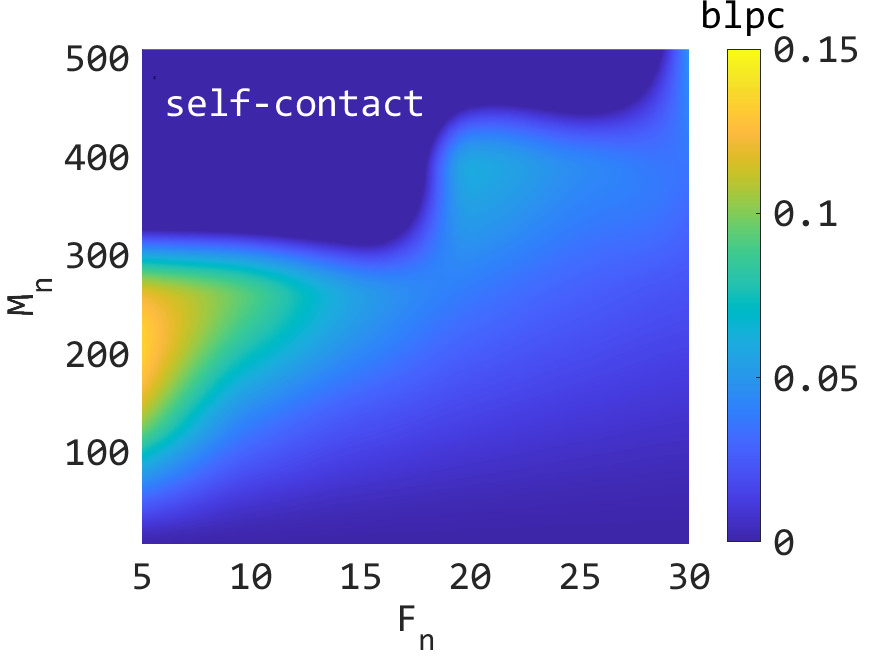}
         \caption{Anguilliform-like}
         \label{fig:Anguilliform-like}
     \end{subfigure}
     \hfill
     \begin{subfigure}[b]{0.49\textwidth}
         \centering
         \includegraphics[scale=0.45]{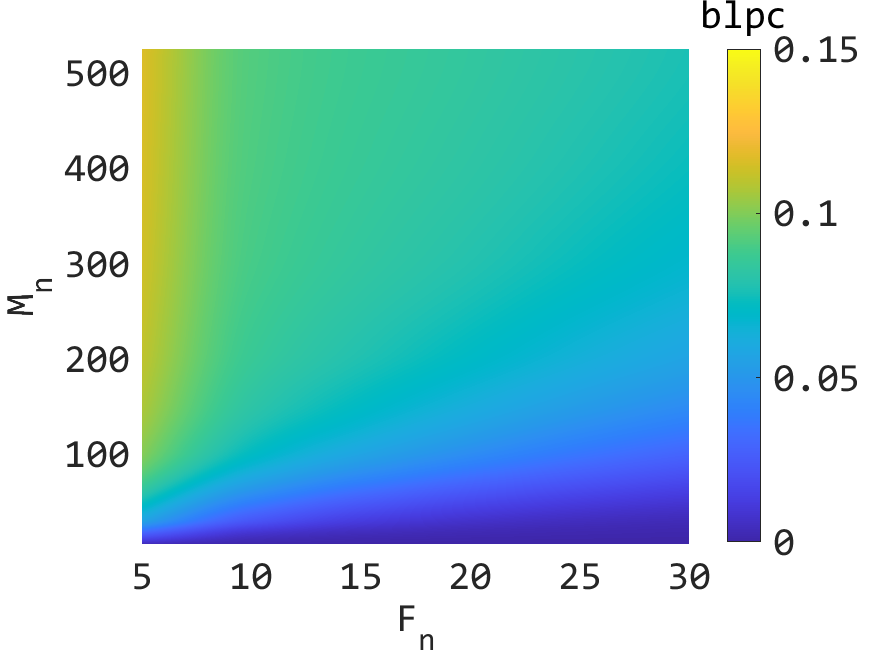}
         \caption{Carangiform-like}
         \label{fig:Carangiform-like}
     \end{subfigure}
     \hfill
     \begin{subfigure}[b]{0.49\textwidth}
         \centering
         \includegraphics[scale=0.45]{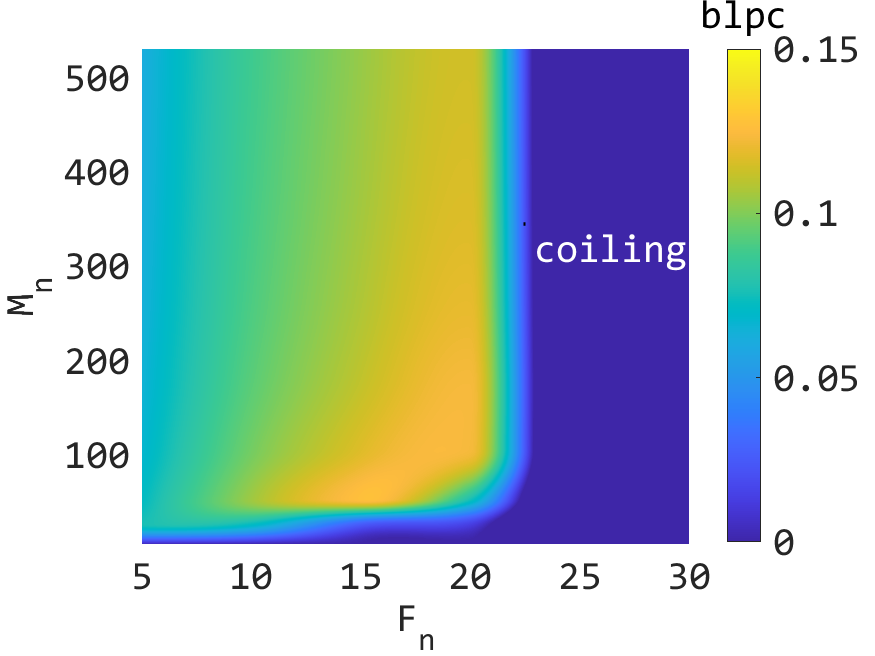}
         \caption{Drag-induced}
         \label{fig:Drag-induced}
     \end{subfigure}
     \begin{subfigure}[b]{0.49\textwidth}
         \centering
         \includegraphics[scale=0.45]{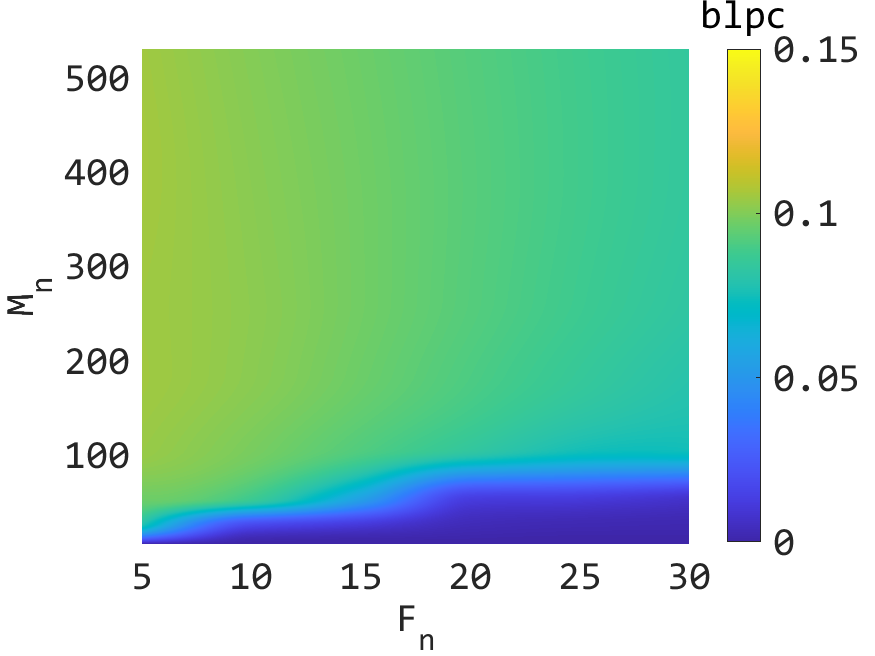}
         \caption{Field-induced}
         \label{fig:Field-induced}
     \end{subfigure}
     \begin{subfigure}[b]{0.49\textwidth}
         \centering
         \includegraphics[scale=0.45]{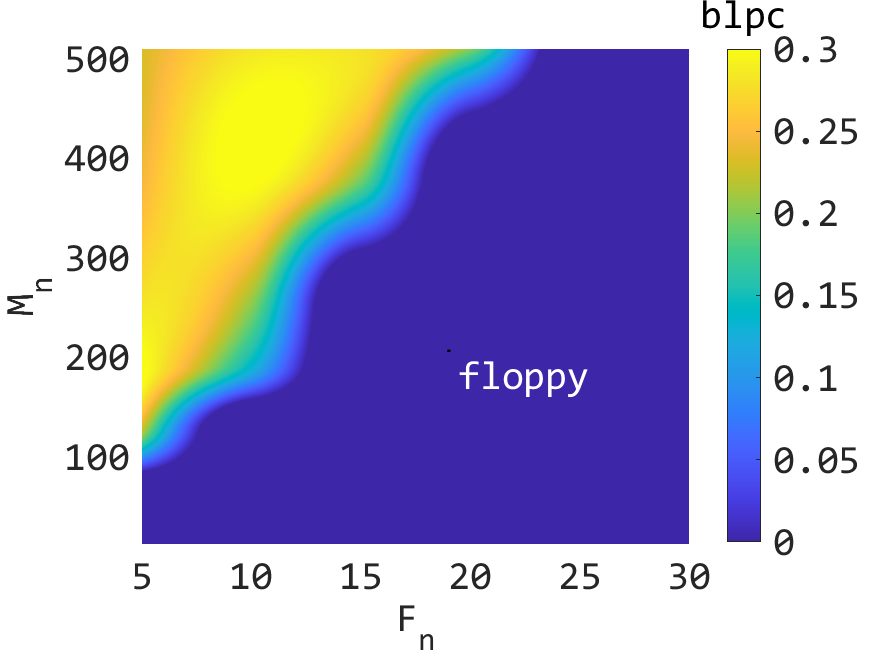}
         \caption{Finger-shaped}
         \label{fig:Finger-shaped}
     \end{subfigure}
        \caption{Variation of blpc (color maps) with respect to M$_\text{n}$ (ranging from 0 to 500) and F$_\text{n}$ (ranging from 5-30) for different MSRS. While the values of $L_\text{0}/L$ are: (a) 1 (fully active), (b) 0.55 (partially active). (c) 0.33 (partially active), (d) 0.33 (partially active), and (e) 1 (fully active), $L/W$ has values of: (a) 8, (b) 1, (c) 5.77, (d) 5.77, and (e) 1. For movies of a selection of these swimmers, see the Supplementary Information.}
        \label{fig:default}
\end{figure*}

\subsubsection{Effect of M$_\text{n}$ and F$_\text{n}$}
We vary the surrounding fluid viscosity so that it manifests itself in F$_\text{n}$ ranging from 5 to 30. Next, keeping other parameters constant, we vary only the magnetic field so as to span the range of M$_\text{n}$ for different values of viscosity. Note that, throughout the paper, we only report the steady-state values and shall explicitly mention them in case of any deviation (e.g., when the swimmer motion is not steady-state or yields negligible net propulsion). Taking a close look at Fig. \ref{fig:default}, we observe that the anguilliform-like (undulatory) MSRS (that has a motion typical of an eel or snake) has a maximum blpc of approximately 0.15; however, when the viscosity is too low (low F$_\text{n}$) and the magnetic field is high (high M$_\text{n}$), the structural deformation is excessive and results in curling (due to high body torques). When this is beyond a limit, different body regions of the swimmer self-contact (see the Supplementary Information for an animation of self-contact), leading to no further swimming. However, when there is a fine balance between M$_\text{n}$ (around 250) and F$_\text{n}$ (around 5), the swimmer displays its best kinematic performance with a forward swimming speed of 0.15 blpc.

In the case of the carangiform-like (undulatory) MSRS that has an oar-like swaying motion (mimicking midge larvae), the maximum blpc is approximately 0.12, and this is when M$_\text{n}$ and F$_\text{n}$ are equal to 500 and 5, respectively. Therefore, at lower viscosities and high magnetic fields, the MSRS is more responsive and follows the magnetic field quite nicely to generate maximum body torques and push the surrounding fluid backward, resulting in enhanced swimming speeds. At higher values of F$_\text{n}$ and lower magnetic fields, the swimmer does not follow (and utilize) the entire span of the directional magnetic field. This results in lower values of blpc. Another important thing to note is that this MSRS is quite stable and does not result in self-contact (like the anguilliform-like MSRS) even if the magnetic field increases. Rather, the swimming performance reaches a plateau (stays constant) with further increase in magnetic field.

The other three MSRSs are all helical: the drag-induced MSRS has one end magnetized, and it follows the rotating magnetic field precisely. However, the rest body region of the swimmer has no magnetization, and magnetic body torques do not act on this region. Nevertheless, there is the presence of the surrounding fluid that imposes drag forces (resistance to motion). As a natural consequence of this FSI, a twisted shape is generated until a steady state is reached \cite{namdeo2014numerical}. Further, the swimmer conforms to this shape and propels like a rigid body. However, when the fluid viscosity is too high, the drag forces also scale up leading to excessive twisting of the swimmer's body; this leads to coiling (see the Supplementary Information for an animation of coiling). It is important to note that the self-contact in anguilliform-like MSRS was a manifestation of excessive bending, while for drag-induced MSRS, it is excessive twisting. However, before this failure limit, this (drag-induced) swimmer has a maximum swimming speed of approximately 0.13 blpc.

While only one end is magnetized for the drag-induced MSRS, both the ends of the field-induced swimmer are magnetized, although in opposite directions (to induce a twisting primarily due to opposite magnetic body torques). Therefore, in principle, the twisting is limited only to one twist per body length. Hence, even when the magnetic fields increase, and irrespective of the fluid viscosity, there is no occurrence of excessive body twisting or coiling. The maximum kinematic performance for the field-induced MSRS is around 0.11, and this swimmer has the innate ability of on-the-fly bi-directional swimming (upon magnetic field direction reversal).

The finger-shaped MSRS has two outer flaps (slender protrusions) that take on a certain shape owing to FSI with the surrounding fluid and magnetic torques (acting upon the entire swimmer body apart from the central flap). The outer flaps rotate around the central flap (that only undergoes pure FSI as it has no magnetization) as a rigid body after reaching the steady-state shape (beyond which there is no structural deformation). The swimmer simply follows the rotating external magnetic field and rotates like a rigid body to swim forward (similar to the typical cork-screw motion). It is crucial to note that although this swimmer has a high swimming speed of around 0.3, this is the case only when a certain minimum magnetic field is reached (for a certain fluid viscosity). Below this limit of M$_\text{n}$, the swimmer does not propel forward as it is too floppy to propel forward (see the Supplementary Information for an animation of a floppy swimmer).

\subsubsection{Effect of normalized magnetic length $L_0/L$}
\begin{figure*}
\captionsetup[subfigure]{justification=centering}
    \centering
     \begin{subfigure}[b]{0.49\textwidth}
         \centering
         \includegraphics[scale=0.45]{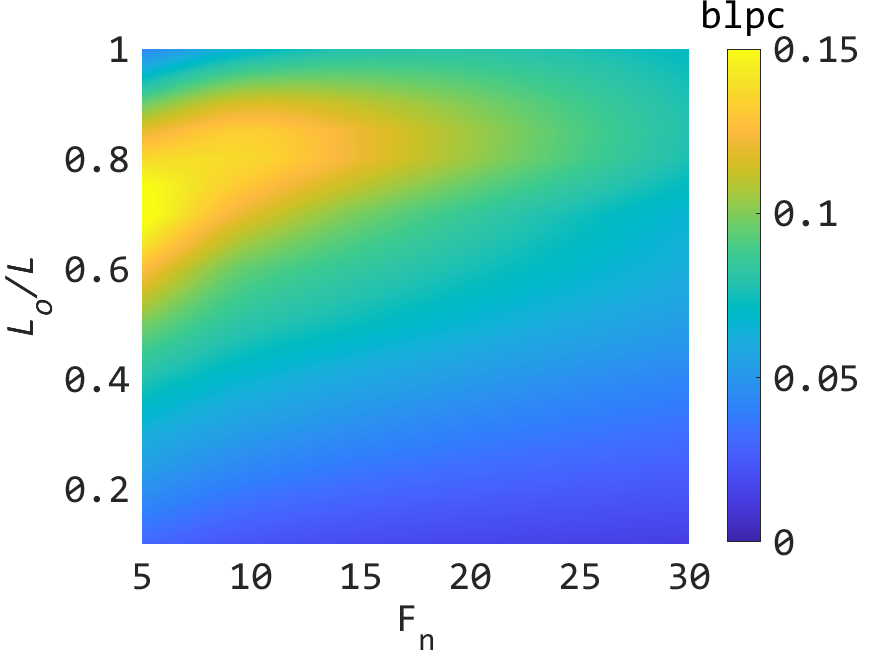}
         \caption{Carangiform-like}
         \label{fig:Carangiform-like}
     \end{subfigure}
     \hfill
     \begin{subfigure}[b]{0.49\textwidth}
         \centering
         \includegraphics[scale=0.45]{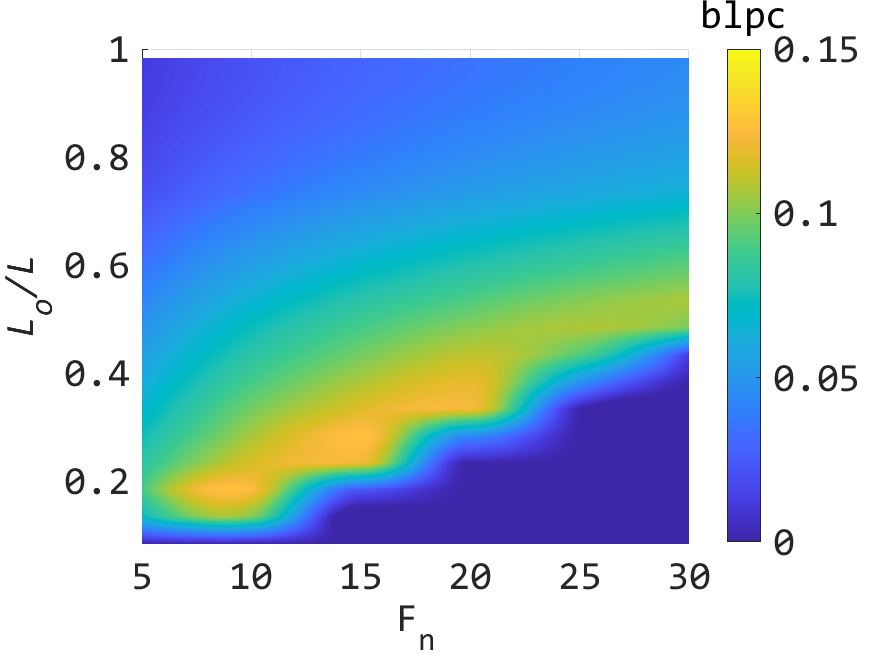}
         \caption{Drag-induced}
         \label{fig:Drag-induced}
     \end{subfigure}
     \begin{subfigure}[b]{0.49\textwidth}
         \centering
         \includegraphics[scale=0.45]{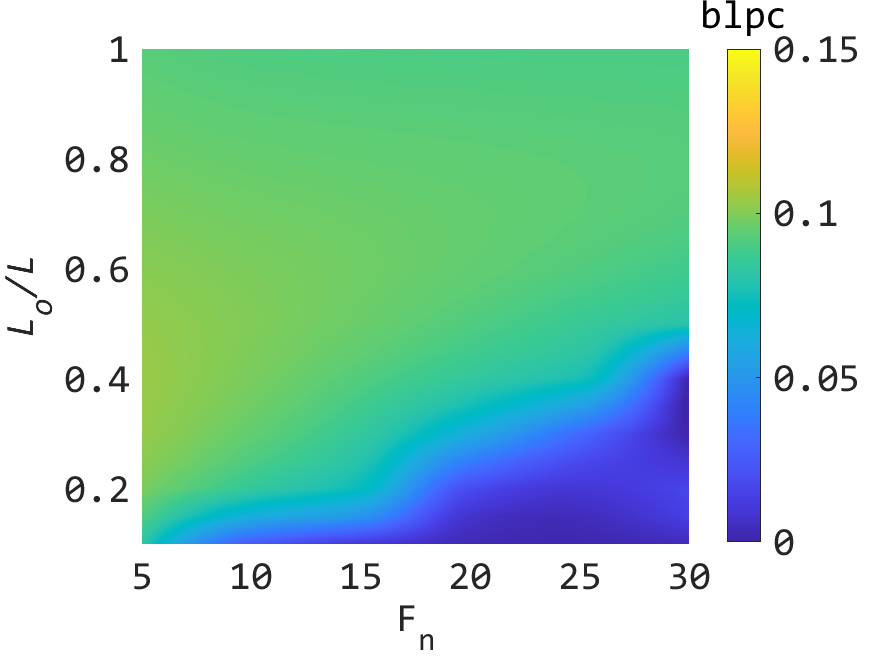}
         \caption{Field-induced}
         \label{fig:Field-induced}
     \end{subfigure}
     \begin{subfigure}[b]{0.49\textwidth}
         \centering
         \includegraphics[scale=0.45]{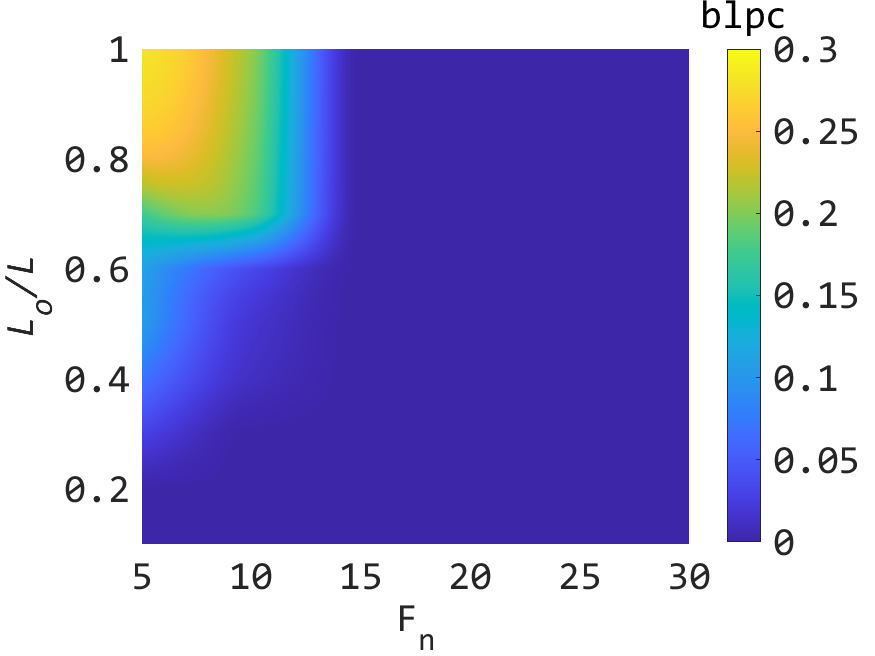}
         \caption{Finger-shaped}
         \label{fig:Finger-shaped}
     \end{subfigure}
        \caption{Variation of blpc (color maps) with respect to $L_\text{0}/L$ (ranging from 0 to 1) and F$_\text{n}$ (ranging from 5-30) for different MSRS. While the amplitude of the magnetic field is held constant at 20mT, $L/W$ has values of: (a) 1, (b) 5.77, (c) 5.77, and (d) 1.}
        \label{fig:magneticlength}
\end{figure*}

Not only does M$_\text{n}$ depend on the magnetic field, but also upon the length of the magnetic portion ($L_\text{0}$) \cite{pramanik2023magnetic}. Therefore, we study the variation of blpc for different values of $L_\text{0}$ (see Fig. \ref{fig:magneticlength}) keeping the magnetic field constant. For different values of fluid viscosity (i.e., F$_\text{n}$), we plot the variation of blpc for different magnetic portions. Please note that we analyze this for all the above-mentioned MSRSs apart from the anguilliform swimmer because this MSRS is entirely magnetized (by its inherent design).

We observe that the carangiform-like swimmer has a maximum blpc around 0.15 for F$_\text{n}$=5 and $L_\text{0}/L$=0.8 (i.e., 80\% magnetized). It is obvious that with increasing viscosity (i.e., F$_\text{n}$), the swimming kinematic performance reduces due to higher viscous resistance and the swimmer's inability to precisely follow the magnetic field. However, for any specific F$_\text{n}$, the swimming speed increases with increasing magnetic portion (until when $L_\text{0}/L$=0.9-1.0, which indicates that the carangiform swimmers behave like a rigid oscillating body that is incapable of achieving any spatial symmetry breaking). Otherwise, with decreasing values of $L_\text{0}/L$, the swimmer pushes more fluid backward to enable enhanced forward swimming motion (thanks to its oar-like swaying).

For the drag-induced MSRS, too high a value of $L_\text{0}/L$ reduces the passive length subjected to pure viscous forces from the surrounding fluid. This is when the magneto-responsive (active) region behaves like a rigid body and rotates synchronously with the magnetic field. Thus, owing to lower effective drag, the twist reduces for this MSRS, resulting in lower swimming speeds. However, when $L_\text{0}/L$ is too low, the drag is too high for the passive region, leading to coiling and self-contact (as explained previously). Therefore, there is an optimum value of $L_\text{0}/L$ for which the blpc is maximum for a specified F$_\text{n}$. The optimal $L_\text{0}/L$ is observed to increase with increasing F$_\text{n}$ because higher viscous forces make the swimmer more vulnerable to coil (due to high fluidic resistance).

The swimming performance of the field-induced MSRS is almost constant throughout the F$_\text{n}$-$L_\text{0}/L$ plane. This is because the chirality is almost fixed due to opposite magnetic body torques at the two ends. With increasing $L_\text{0}/L$, the twist remains (almost) unaltered, and the swimming speed is primarily dependent on the overall fluid resistance felt by the swimmer. Therefore, with increasing values of F$_\text{n}$, the blpc reduces, although for a minimum $L_\text{0}/L$=0.4, the optimal blpc is reached. For this MSRS, there is no coiling or self-contact observed irrespective of F$_\text{n}$ or $L_\text{0}/L$.

The swimming behavior of the finger-shaped MSRS depends significantly on the F$_\text{n}$-$L_\text{0}/L$ plane. At lower values of F$_\text{n}$, higher values of $L_\text{0}/L$ are necessary to propel the swimmer (else, the swimmer is within its floppy regime as discussed earlier). When the swimmer is rather entirely magnetized, net propulsion is observed, thanks to higher fluidic interactions balanced by increased magnetic body torques. However, for higher values of F$_\text{n}$, the swimmer is always within its floppy regime for all values of $L_\text{0}/L$ (for the magnetic field chosen). It is important to mention that if the magnetic field is increased further, keeping $L_\text{0}/L$=1.0, the swimmer would still propel forward (see Fig. \ref{fig:default}e).

\begin{figure*}
\captionsetup[subfigure]{justification=centering}
     \centering
     \begin{subfigure}[b]{0.49\textwidth}
         \centering
         \includegraphics[scale=0.45]{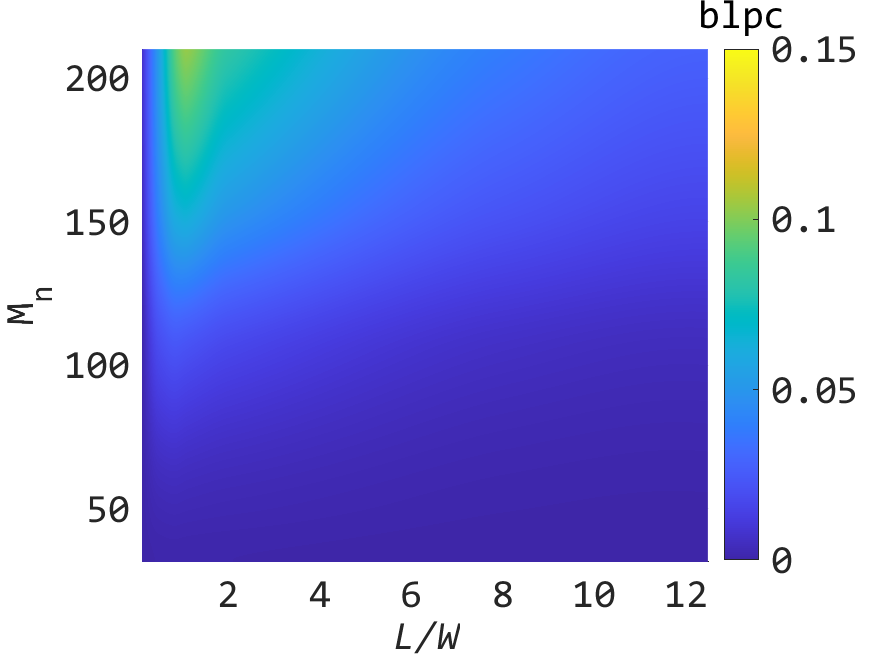}
         \caption{Anguilliform-like}
         \label{fig:Carangiform-like}
     \end{subfigure}
     \hfill
     \begin{subfigure}[b]{0.49\textwidth}
         \centering
         \includegraphics[scale=0.45]{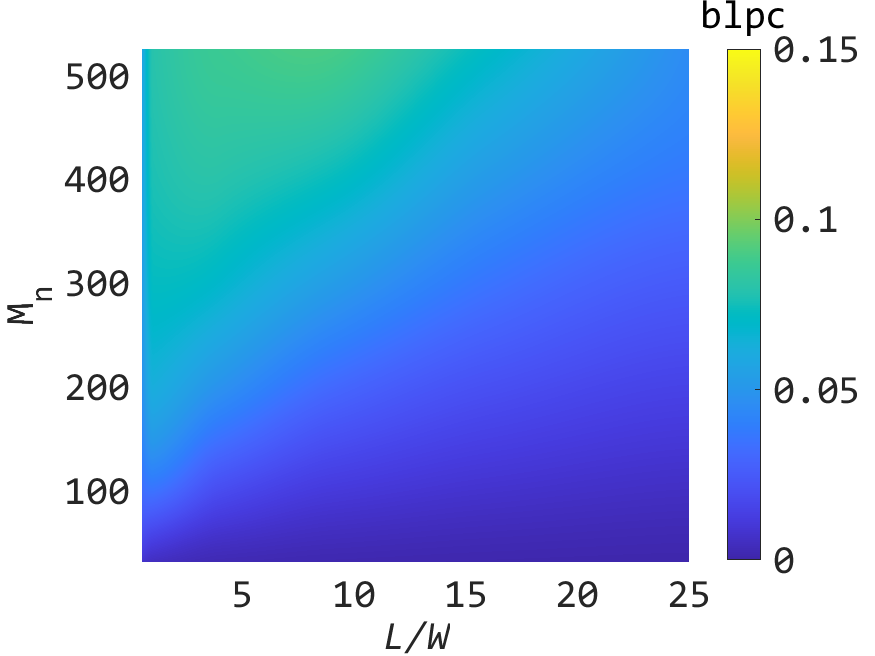}
         \caption{Carangiform-like}
         \label{fig:Anguilliform-like}
     \end{subfigure}
     \begin{subfigure}[b]{0.49\textwidth}
         \centering
         \includegraphics[scale=0.45]{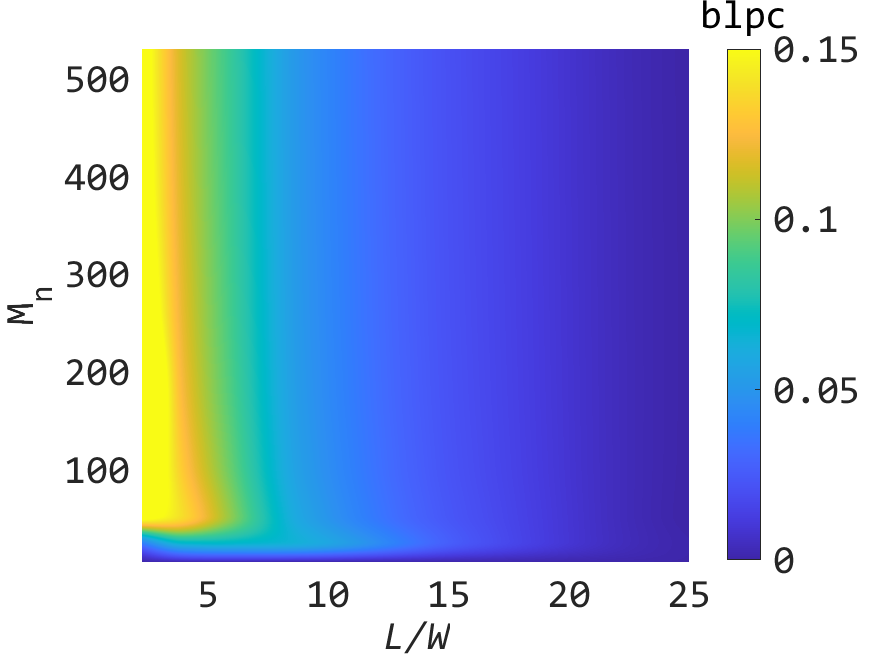}
         \caption{Drag-induced}
         \label{fig:Drag-induced}
     \end{subfigure}
     \begin{subfigure}[b]{0.49\textwidth}
         \centering
         \includegraphics[scale=0.45]{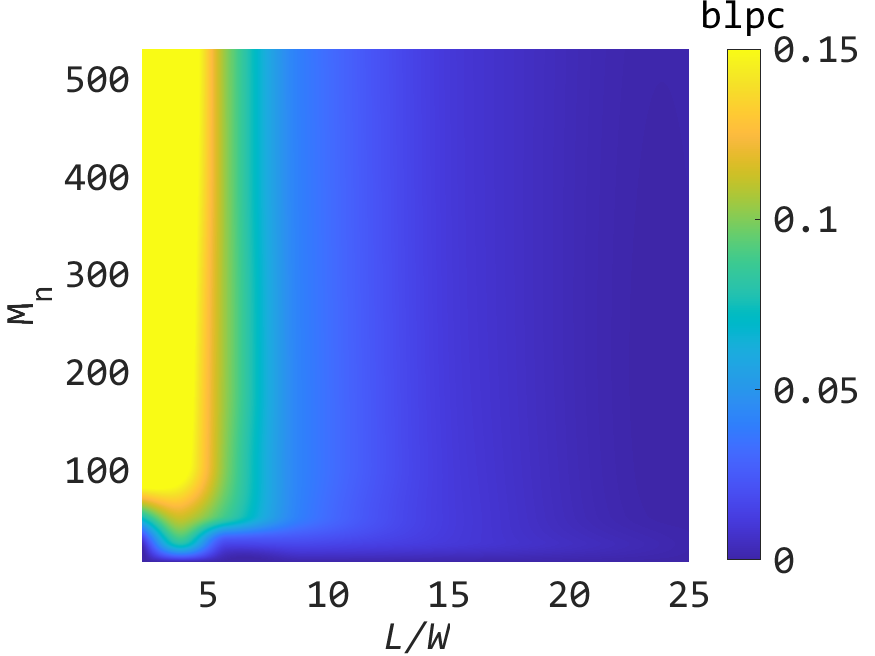}
         \caption{Field-induced}
         \label{fig:Field-induced}
     \end{subfigure}
        \caption{Variation of blpc (color maps) with respect to M$_\text{n}$ and $L/W$ for different MSRS. While the values of $L_\text{0}/L$ are: (a) 1 (fully active), (b) 0.55 (partially active). (c) 0.33 (partially active), and (d) 0.33 (partially active), F$_\text{n}$ has a constant value of 10.}
        \label{fig:aspectratio}
\end{figure*}

\subsubsection{Effect of aspect ratio $L/W$}
Not only do the magnetics and fluid dynamics influence the swimming performance of these MSRSs, but also the geometry (particularly, the aspect ratio) plays a crucial role (see Fig. \ref{fig:aspectratio}). These soft robotic swimmers experience two-way fully coupled FSI and their geometry affects the surrounding fluidic manipulation in generating the thrust required for swimming. The carangiform swimmer has the highest $blpc$ when it has a square geometry. But, when the length increases (and the width decreases accordingly to keep the area, i.e., material volume, the same), the swimming speed reduces. This is because the fluid pushed backward effectively reduces due to different FSI going on. The increase in magnetic field simply increases the responsiveness of the swimmer, and higher fluid quantities are displaced to generate more thrust. Hence, in this study, we consider the carangiform-like undulatory swimmer to have an aspect ratio equal to one (square geometry). Similar reasoning can, in principle, be applied to the anguilliform-like swimmer because of similar undulatory motion (bending of the swimmer's body to displace the surrounding fluid).

However, the situation is different for field-induced and drag-induced swimmers. Twisting is naturally difficult when the swimmer aspect ratio decreases. When we set the aspect ratio equal to one, the swimmers do not twist properly. As a consequence, they do not demonstrate swimming and net forward propulsion. With higher aspect ratios, it starts becoming easier to twist these MSRSs. Hence, a square swimmer is the least desired, while one with a higher aspect ratio of around 5 exhibits optimal swimming kinematics. But then, when the aspect ratio is increased even further, although the twist becomes easier, the effective FSI that essentially displaces the fluid for forward propulsion diminishes, generating a lower thrust that manifests in slow swimming speeds. At very high aspect ratios, we observe no net propulsion although the twisting is easily achievable. Finally, when we varied the aspect ratio for the finger-shaped swimmer, we found that other designs were always within the floppy regime. Hence, we do not mention this in Fig. \ref{fig:aspectratio}.

\begin{figure*}[h]
\captionsetup[subfigure]{justification=centering}
     \centering
     \begin{subfigure}[b]{0.99\textwidth}
         \centering
         \includegraphics[scale=0.3]{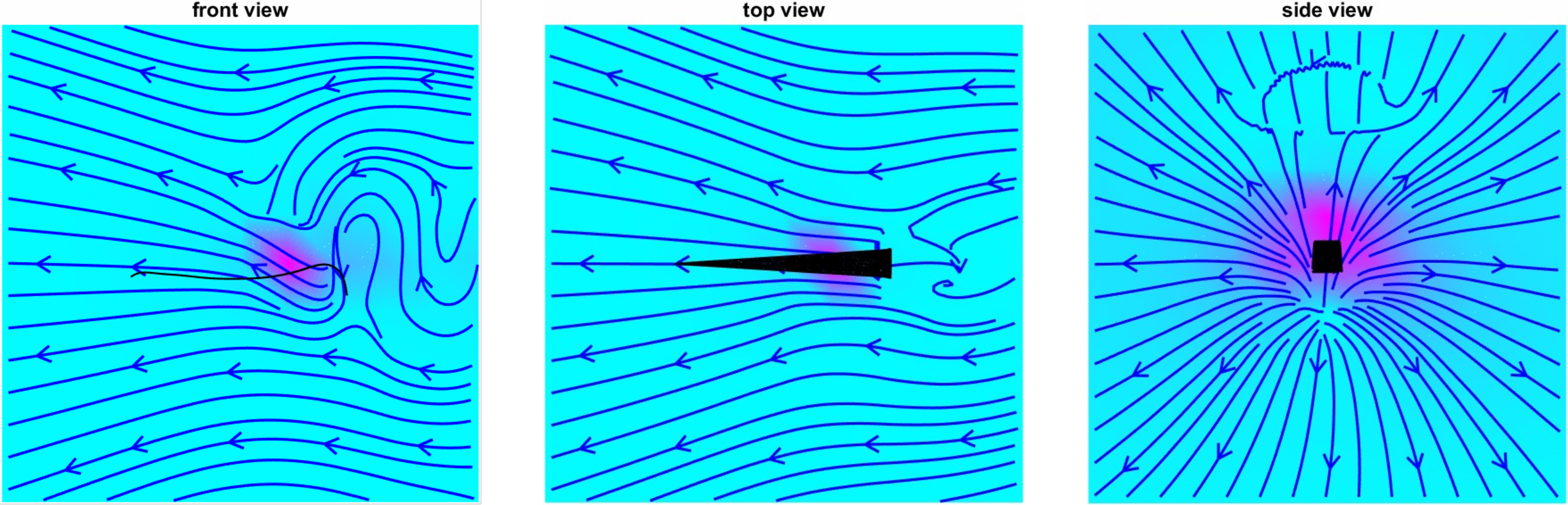}
         \caption{Flow field around the anguilliform-like undulatory MSRS.}
         \label{fig:Anguilliform-like}
     \end{subfigure}
     \hfill
    \begin{subfigure}[b]{0.99\textwidth}
         \centering
         \includegraphics[scale=0.3]{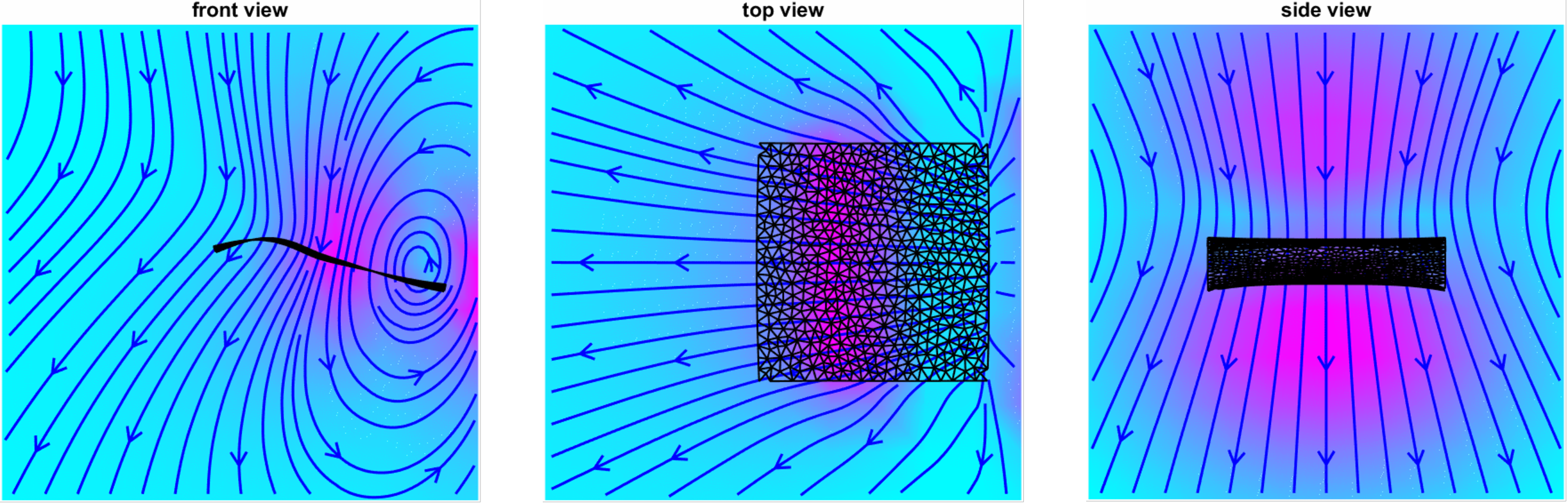}
         \caption{Flow field around the carangiform-like undulatory MSRS.}
         \label{fig:Carangiform-like}
     \end{subfigure}
     \hfill
     \begin{subfigure}[b]{0.99\textwidth}
         \centering
         \includegraphics[scale=0.3]{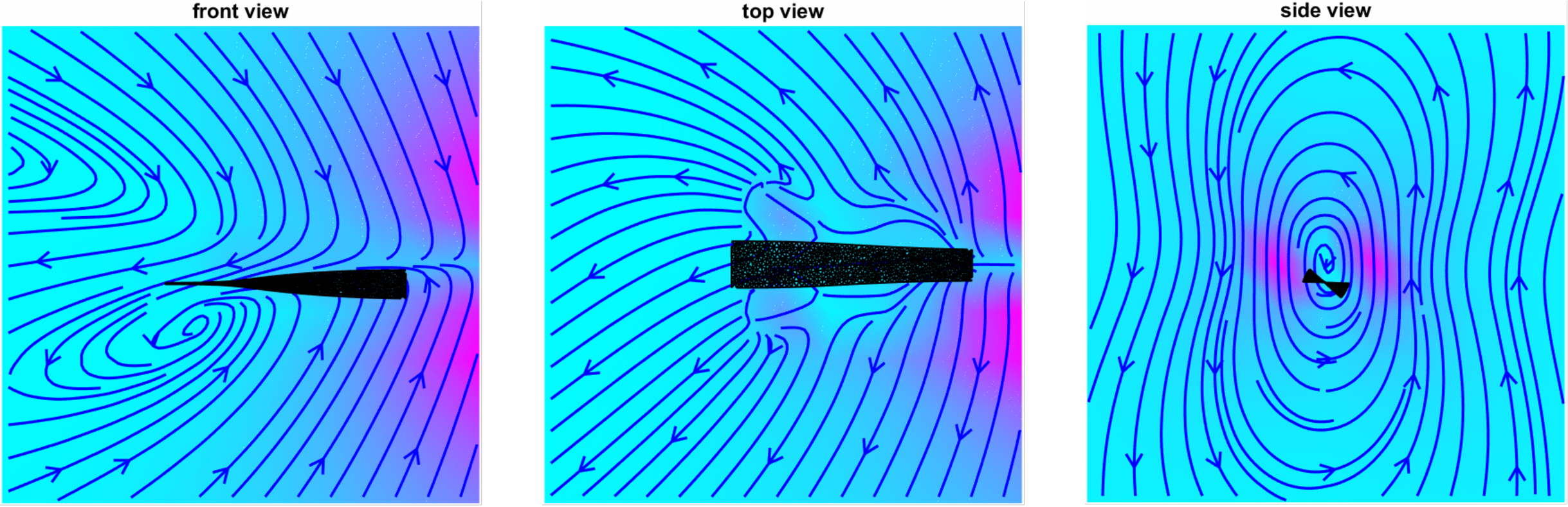}
         \caption{Flow field around the (field- or) drag-induced helical MSRS.}
         \label{fig:Dragfield-induced}
     \end{subfigure}
     \hfill
     \begin{subfigure}[b]{0.99\textwidth}
         \centering
         \includegraphics[scale=0.3]{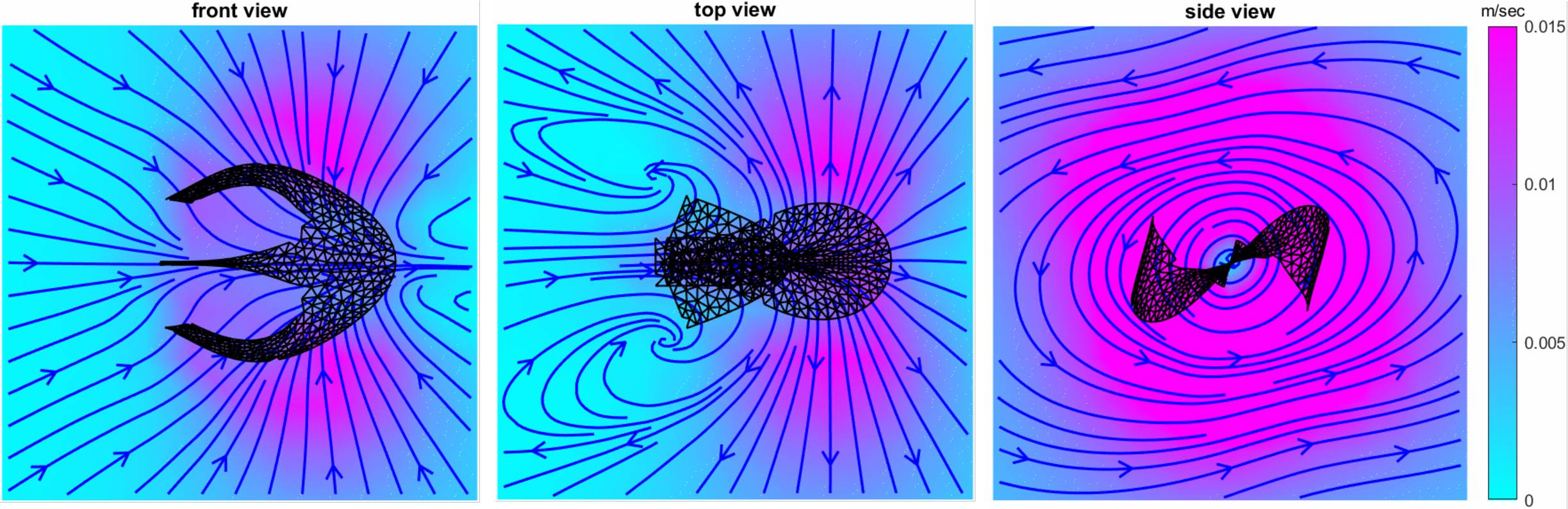}
         \caption{Flow field around the finger-shaped helical MSRS.}
         \label{fig:Finger-shaped}
     \end{subfigure}
     \hfill
        \caption{Snapshots of the flow fields around the MSRS: \textbf{front} (x-z), \textbf{top} (x-y), and \textbf{side} (y-z) views. The blue lines (with arrows) represent the streamlines. For all the views, we plot the \textbf{magnitude} of the fluid velocity obtained from the computational fluid dynamics model using the Green's function approximation. All the swimmers propel along the +ve x-direction as mentioned in Fig. \ref{fig:swimmers}.}
        \label{fig:flowfield}
\end{figure*}

\subsection{Flow fields}
Next, we study the evolution of the surrounding flow field for each MSRS to understand the swimming behavior in a quantitative and precise manner (see Fig. \ref{fig:flowfield}). Specifically, we calculate the volume flowrate, \textit{Q} (defined as the average fluid velocity, \textit{v} multiplied by the square of the characteristic body length L) of the surrounding fluid in all three directions: x, y, and z (in the Cartesian coordinate system) during one complete swimming cycle (as a function of time). Additionally, we report the time-average flowrate $\Bar{Q}$ and correlate it with blpc for each swimmer. Please note that the flowrate(s) and time have been normalized with respect to $\Bar{L}^3/T$ and \textit{T}, respectively (and are non-dimensional). The animations can be found in the Supplementary Information (refer to the presentation on flow fields around different swimmers), the time-instant and average flow field velocities are plotted below for one complete swimming cycle (see Fig. \ref{fig:flowrates}). Please note that the direction of propulsion for all the swimmers is along the x-axis (refer Fig. \ref{fig:swimmers}).

For the anguilliform-like undulatory MSRS, the bending structural body deformations are along the x-z plane, and there is hardly any deformation along the y-axis. Therefore, the flowrate along the y-axis is nearly zero, while it is quite higher for the z-axis. The bending deformation occurs in a sinusoidal manner, and this pushes the surrounding fluid alike. Although the variation of the x-flowrate has the same time period as the swimmer motion, the time period of the z-flowrate variation is double. This is clearly due to the traveling bending waves and sinusoidal deformation of the swimmer. Furthermore, despite the variations being quite large, the average flowrates are not too high. For the carangiform-like MSRS, a similar structural (bending) deformation is observed, and the surrounding flow fields are quite similar to that of the anguilliform-like MSRS (with a slight phase shift).

\begin{figure*}
\captionsetup[subfigure]{justification=centering}
     \centering
     \begin{subfigure}[b]{0.49\textwidth}
         \centering
         \includegraphics[scale=0.45]{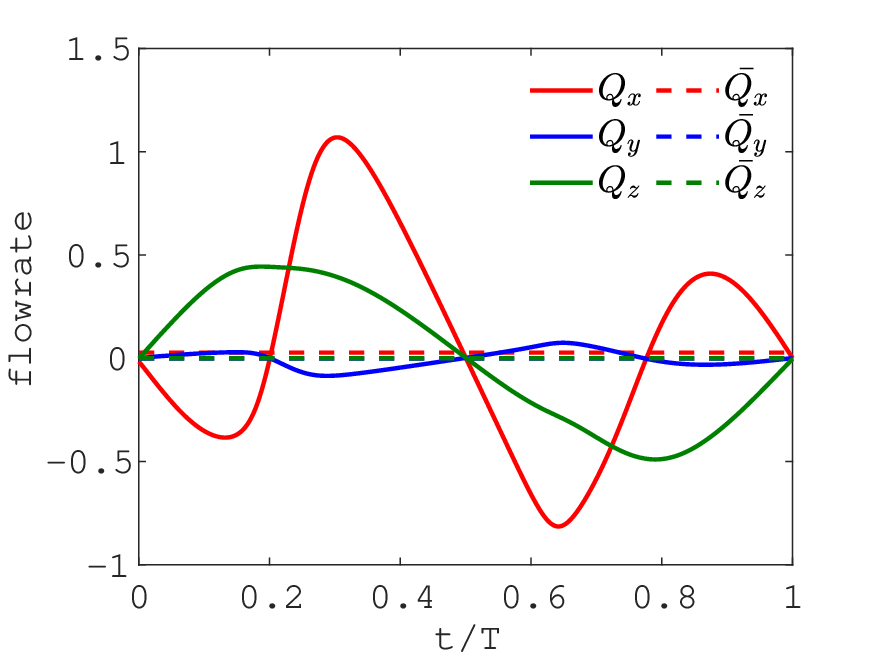}
         \caption{Anguilliform-like}
         \label{fig:Anguilliform-like}
     \end{subfigure}
     \hfill
     \begin{subfigure}[b]{0.49\textwidth}
         \centering
         \includegraphics[scale=0.45]{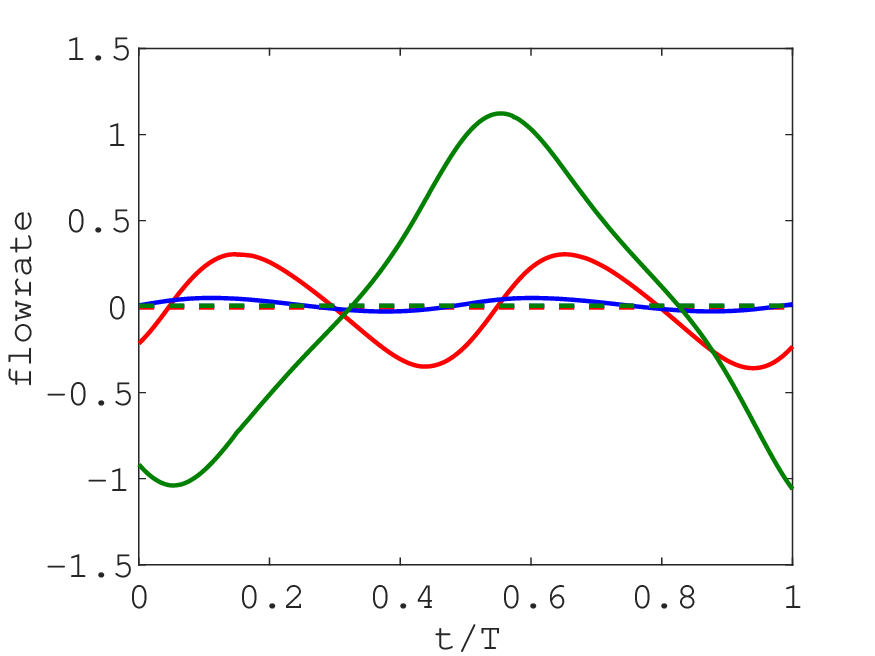}
         \caption{Carangiform-like}
         \label{fig:Carangiform-like}
     \end{subfigure}
     \hfill
     \begin{subfigure}[b]{0.49\textwidth}
         \centering
         \includegraphics[scale=0.45]{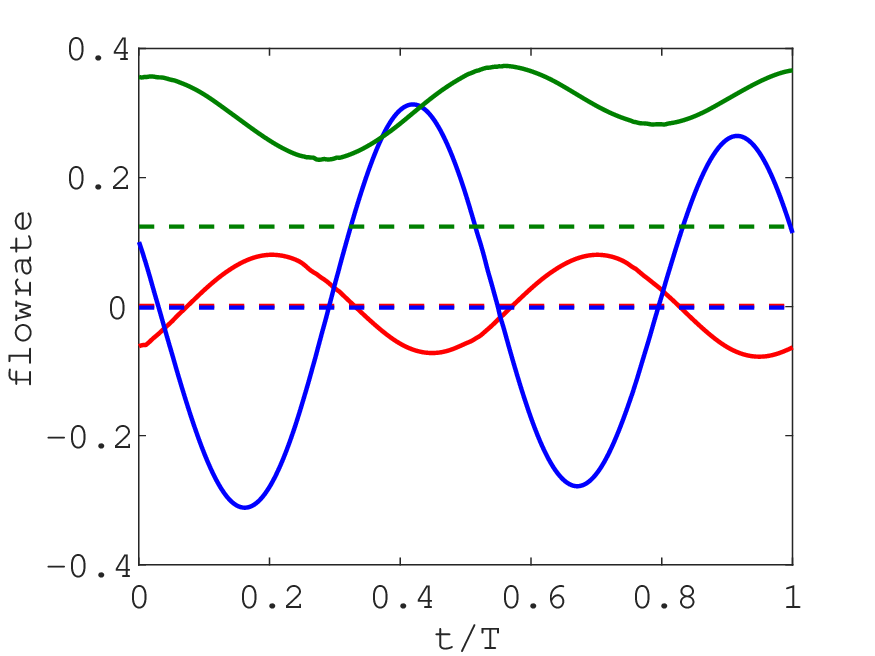}
         \caption{Drag-induced}
         \label{fig:Drag-induced}
     \end{subfigure}
     \begin{subfigure}[b]{0.49\textwidth}
         \centering
         \includegraphics[scale=0.45]{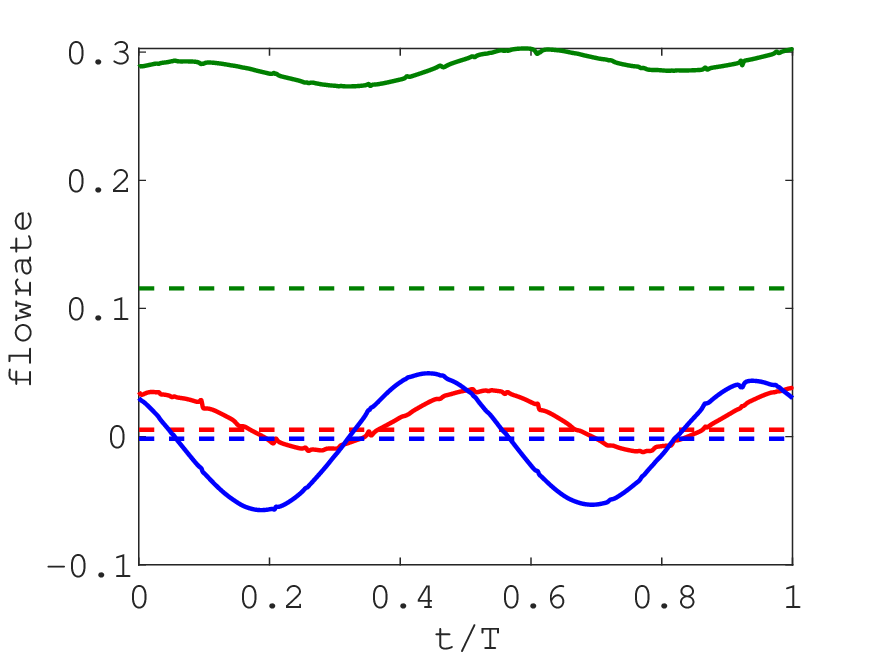}
         \caption{Field-induced}
         \label{fig:Field-induced}
     \end{subfigure}
     \begin{subfigure}[b]{0.49\textwidth}
         \centering
         \includegraphics[scale=0.45]{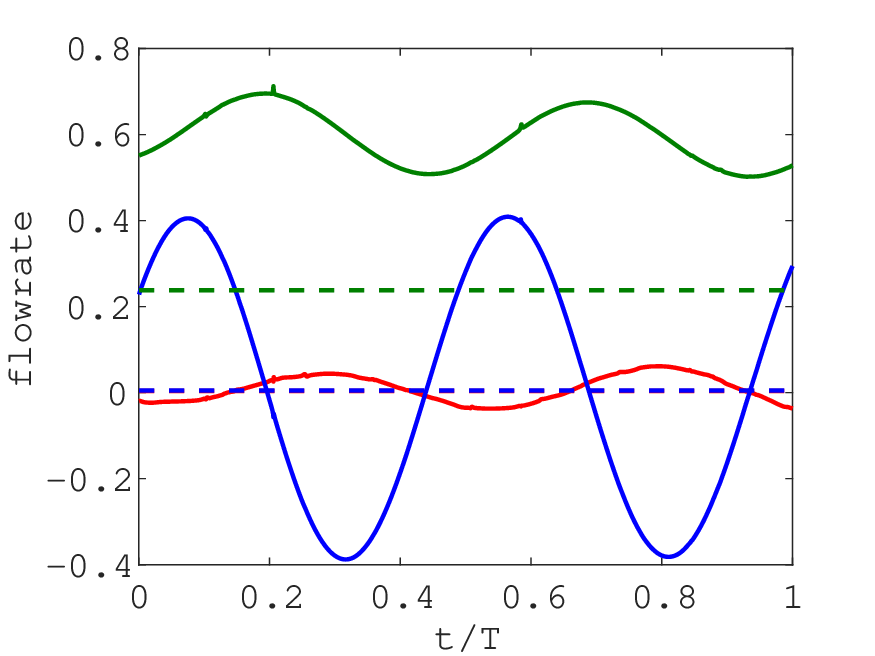}
         \caption{Finger-shaped}
         \label{fig:Finger-shaped}
     \end{subfigure}
     \begin{subfigure}[b]{0.49\textwidth}
         \centering
         \includegraphics[scale=0.45]{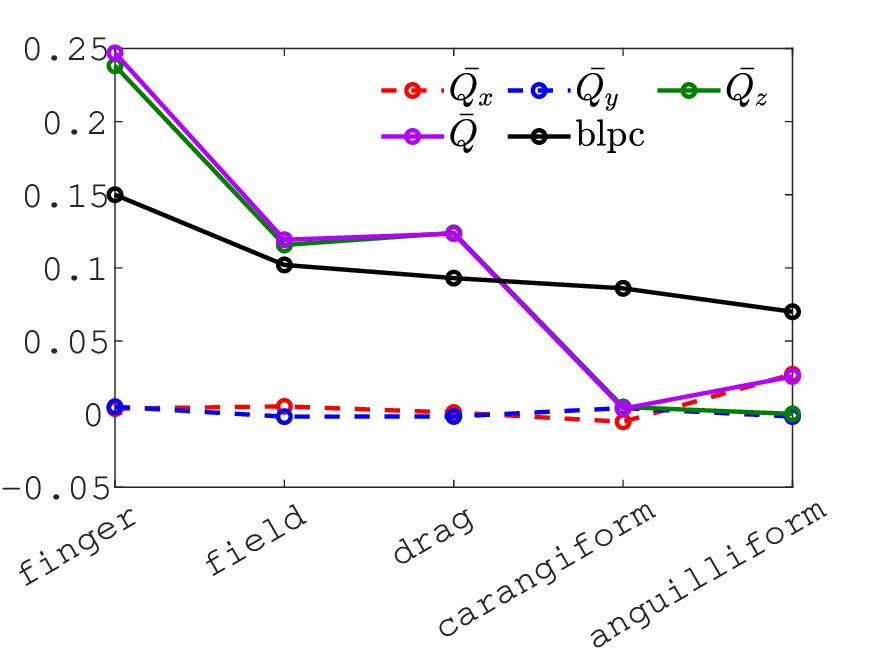}
         \caption{Comparing flowrates and blpc (M$_\text{n}$=200, F$_\text{n}$=10).}
         \label{fig:flowrates}
     \end{subfigure}
        \caption{Time-dependent variation of normalized flowrate(s) as a function of $t/T$ for different swimmers during one complete swimming cycle.}
        \label{fig:flowrates}
\end{figure*}

However, for the drag-induced MSRS, the flowrates also vary sinusoidally. We observe that the x-flowrate and the y-flowrate averages much lower compared to the z-flowrate. The flow fields are quite alike for the field-induced and finger-shaped MSRS. We note that the z-flowrate in all the helical swimmers is significantly higher than either the x-flowrate or y-flowrate. To put perspectives quantitatively for ease of analysis, we plot the respective average flowrates along all axes (and add them up all to count for the total flowrate) along with their respective values of blpc. Interestingly, we observe a strong correlation between the total flowrate ($Q$) and swimming speed (blpc). This supports the argument that the swimmers propel faster when they interact more with the surrounding fluid higher and also displaces them. This is in fine accordance with previous studies carried out for miniaturized ciliary structures displacing the surrounding fluid for net fluidic transportation \cite{khaderi2009nature} or even propulsion \cite{golestanian2008analytic}.

\subsection{Bi-directionality}
Until now, we considered only the forward swimming (and kinematic performance) of different MSRSs based on different non-dimensional numbers that fully incorporate all system parameters. However, during practical applications, it is beneficial if a swimmer has the ability of on-the-fly swimming direction reversal (i.e., bi-directional swimming). Therefore, it is crucial to check the maneuverability of a swimmer.

We assign the nice feature of bi-directionality only if the swimmer undergoes an on-the-fly swimming direction reversal. In the case of the anguilliform swimmer, this is not immediately possible with the vertically oscillating field (as proposed in Fig. \ref{fig:swimmers}) because the magnetic field has no directionality. However, if the externally applied field is rotated so as to align with the (remnant) magnetization direction, and then further rotated in one go, the swimmer would simply follow the field and reorient itself along the opposite direction. Upon further subject to the oscillating field as in Fig. \ref{fig:swimmers}, it would swim along the negative x-axis. For the carangiform-like undulatory swimmer, the oscillating magnetic field is directional (see Fig. \ref{fig:swimmers}d). Therefore, the swimmer ought to be rotated at the first instance, and then upon subject to external directional field (in the reverse direction), it would traverse along the negative x-axis.

As for the helical swimmers, the drag-induced MSRS and the finger-shaped MSRS do not possess on-the-fly swimming direction reversal when the external rotating magnetic field is reversed. By nature of their design (and directionality of remnant magnetization), the swimmers can propel only along the positive x-axis. Even when the external magnetic rotation direction is reversed, the swimming direction does not change since the swimmers still push the fluid to the left under its steady-state shape generated as a natural consequence of FSI. As an exception, the field-induced MSRS can instantaneously reverse the swimming direction when the external rotational field direction is reversed. This is simply because the remnant magnetization profile of the swimmer is not directional and that the chirality remains fixed (frozen-in by the field) so that only the rotation direction reverses and thus the swimming direction \cite{namdeo2014numerical}.

\subsection{Stability}
\begin{figure*}
\captionsetup[subfigure]{justification=centering}
     \centering
     \begin{subfigure}[b]{0.33\textwidth}
         \centering
         \includegraphics[scale=0.3]{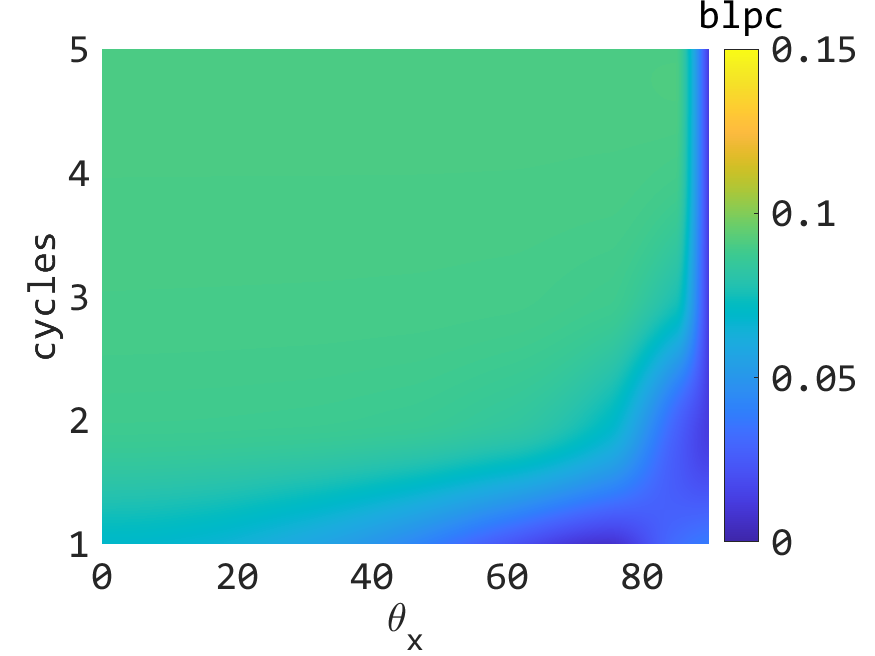}
         \caption{Carangiform-like MSRS: $\theta_\text{x}$}
         \label{fig:stabilitycarangiformx}
     \end{subfigure}
     \begin{subfigure}[b]{0.33\textwidth}
         \centering
         \includegraphics[scale=0.3]{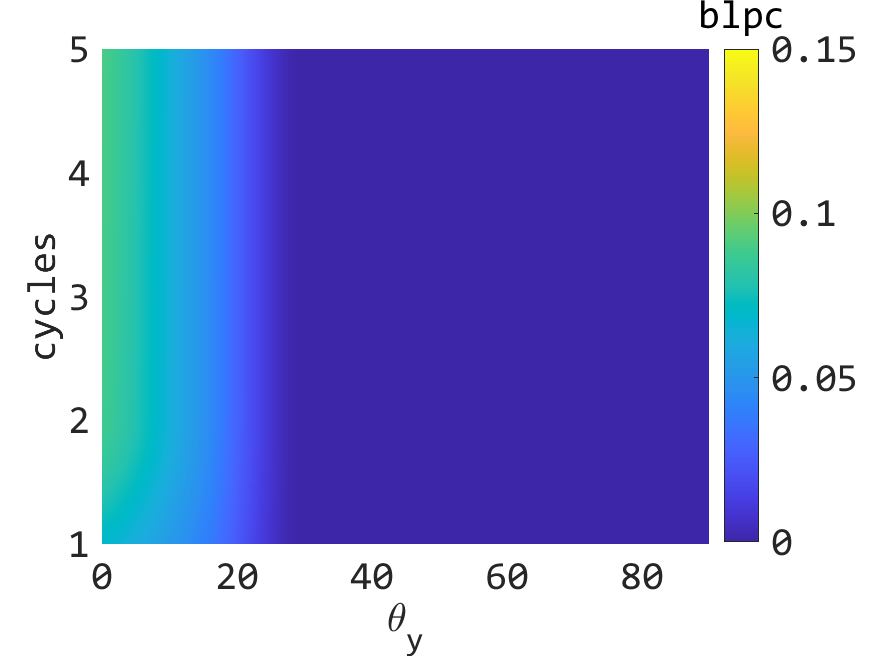}
         \caption{Carangiform-like MSRS: $\theta_\text{y}$}
         \label{fig:stabilitycarangiformy}
     \end{subfigure}
     \begin{subfigure}[b]{0.33\textwidth}
         \centering
         \includegraphics[scale=0.3]{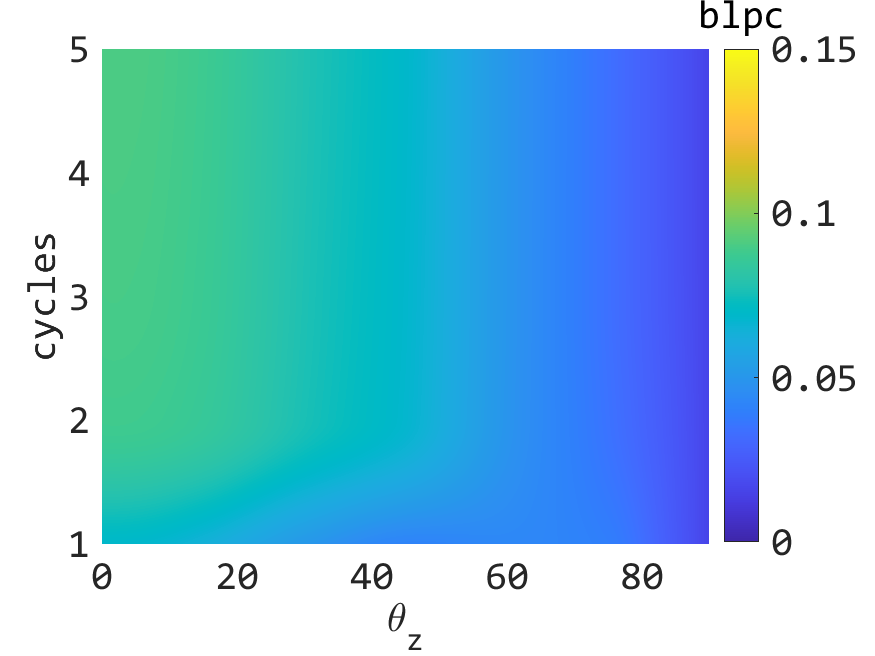}
         \caption{Carangiform-like MSRS: $\theta_\text{z}$}
         \label{fig:stabilitycarangiformz}
     \end{subfigure}
          \begin{subfigure}[b]{0.33\textwidth}
         \centering
         \includegraphics[scale=0.3]{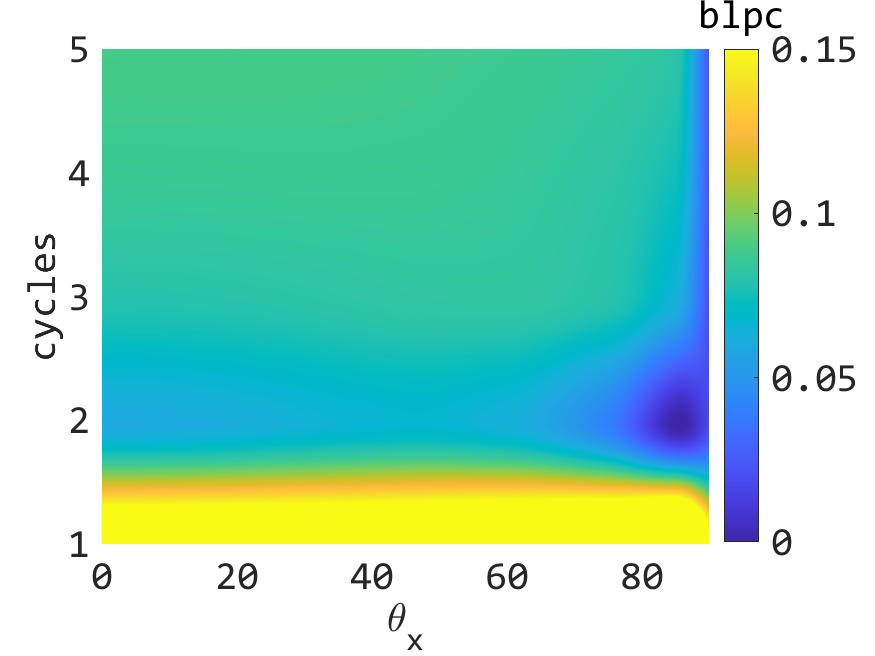}
         \caption{Anguilliform-like MSRS: $\theta_\text{x}$}
         \label{fig:stabilityanguilliformx}
     \end{subfigure}
     \begin{subfigure}[b]{0.33\textwidth}
         \centering
         \includegraphics[scale=0.3]{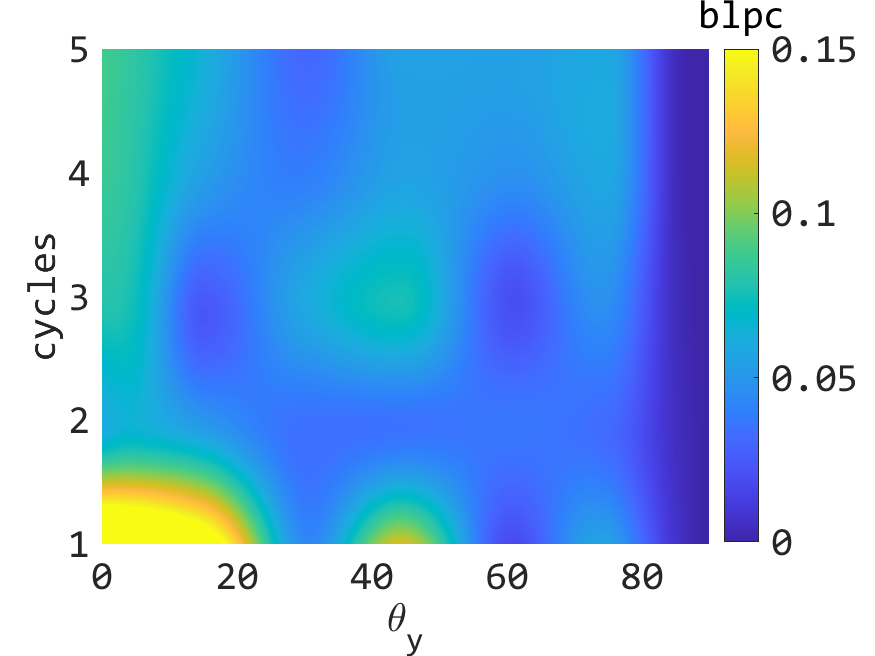}
         \caption{Anguilliform-like MSRS: $\theta_\text{y}$}
         \label{fig:stabilityanguilliformy}
     \end{subfigure}
     \begin{subfigure}[b]{0.33\textwidth}
         \centering
         \includegraphics[scale=0.3]{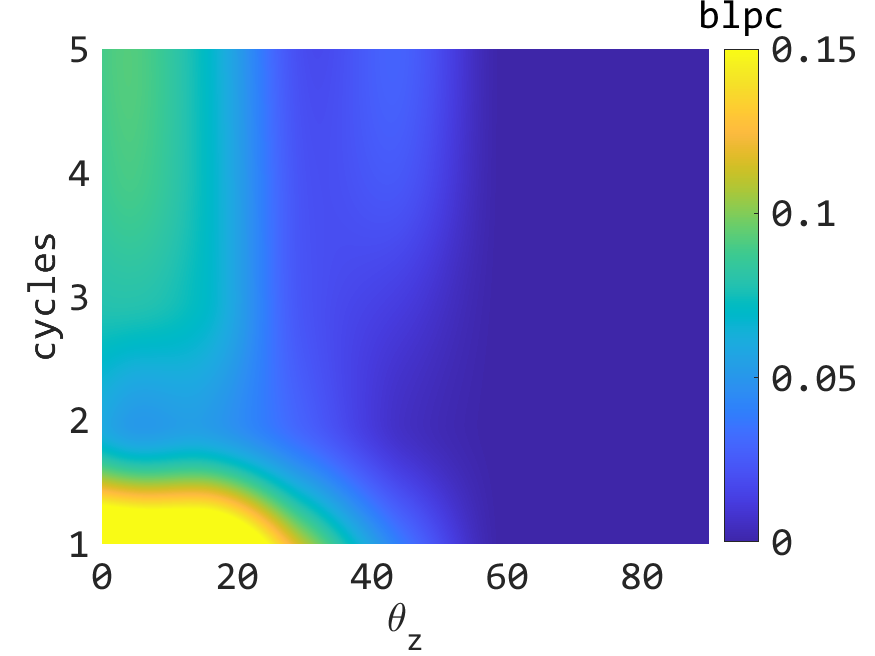}
         \caption{Anguilliform-like MSRS: $\theta_\text{z}$}
         \label{fig:stabilityanguilliformz}
     \end{subfigure}
     \begin{subfigure}[b]{0.33\textwidth}
         \centering
         \includegraphics[scale=0.3]{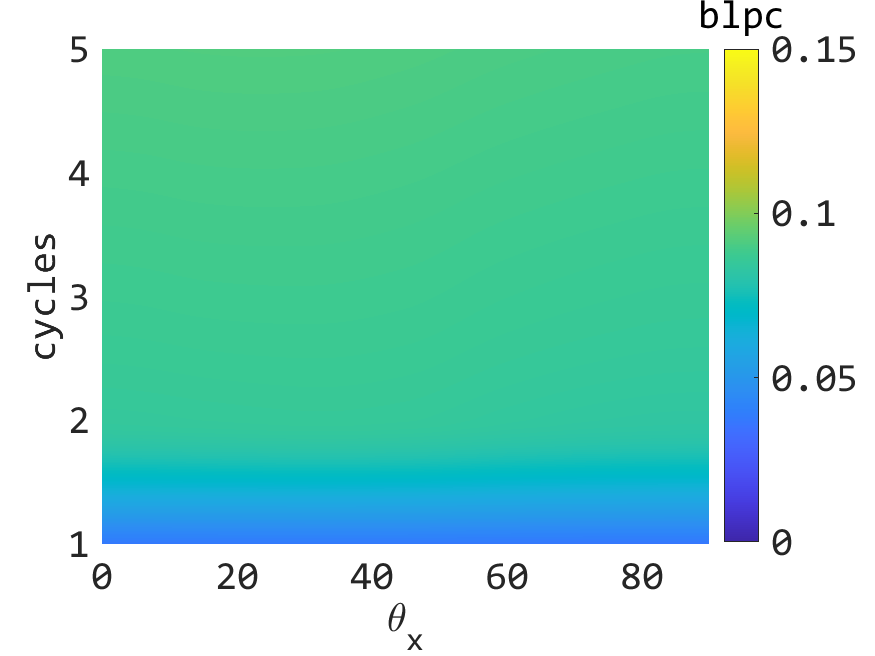}
         \caption{Drag-induced MSRS: $\theta_\text{x}$}
         \label{fig:stabilitydragx}
     \end{subfigure}
     \begin{subfigure}[b]{0.33\textwidth}
         \centering
         \includegraphics[scale=0.3]{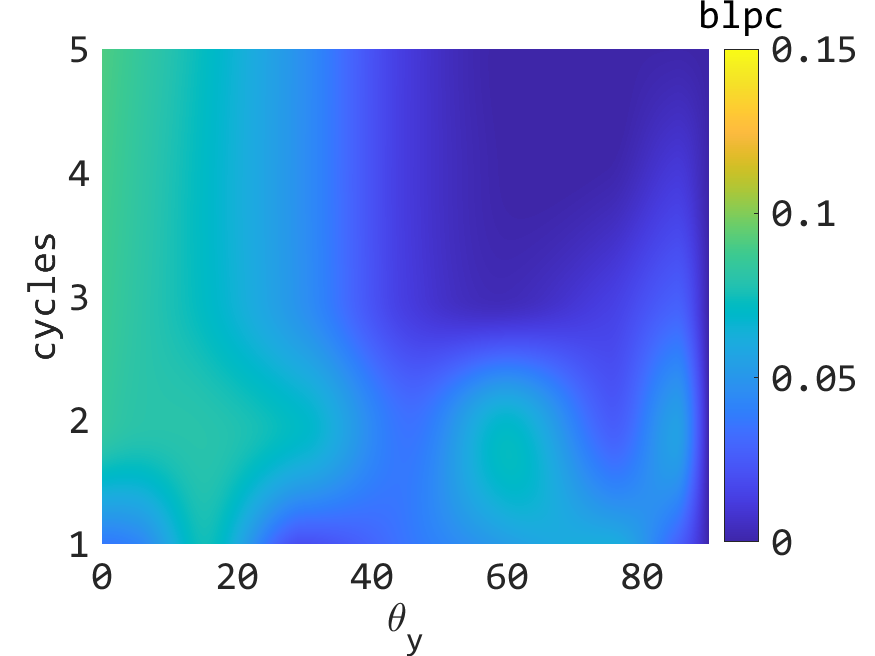}
         \caption{Drag-induced MSRS: $\theta_\text{y}$}
         \label{fig:stabilitydragy}
     \end{subfigure}
     \begin{subfigure}[b]{0.33\textwidth}
         \centering
         \includegraphics[scale=0.3]{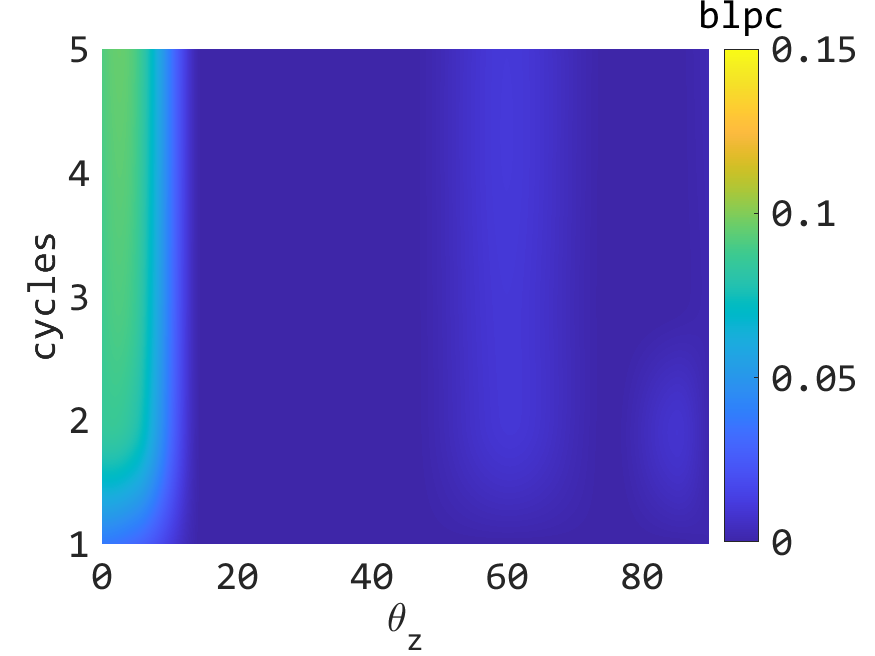}
         \caption{Drag-induced MSRS: $\theta_\text{z}$}
         \label{fig:stabilitydragz}
     \end{subfigure}
     \begin{subfigure}[b]{0.33\textwidth}
         \centering
         \includegraphics[scale=0.3]{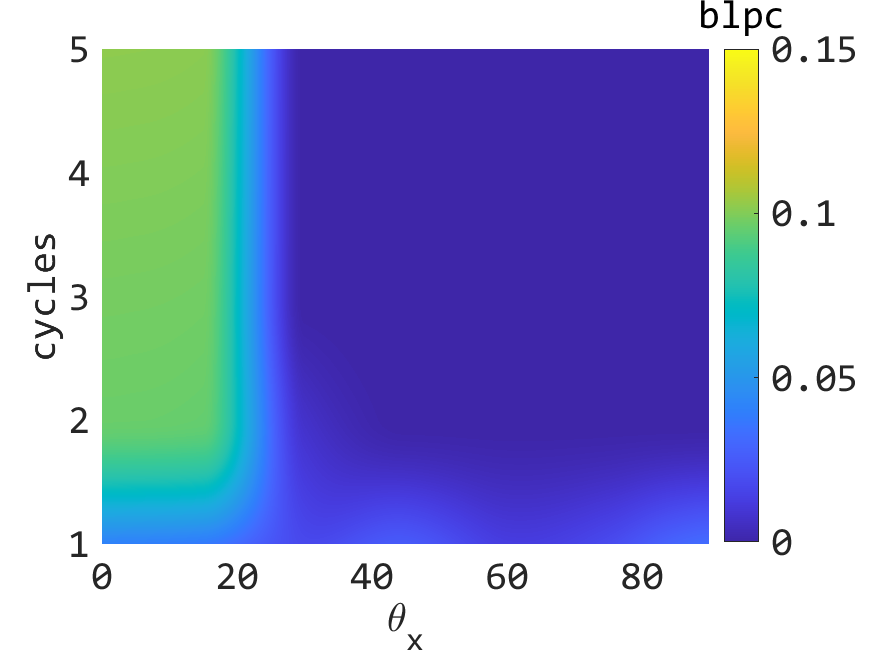}
         \caption{Field-induced MSRS: $\theta_\text{x}$}
         \label{fig:stabilityfieldx}
     \end{subfigure}
     \begin{subfigure}[b]{0.33\textwidth}
         \centering
         \includegraphics[scale=0.3]{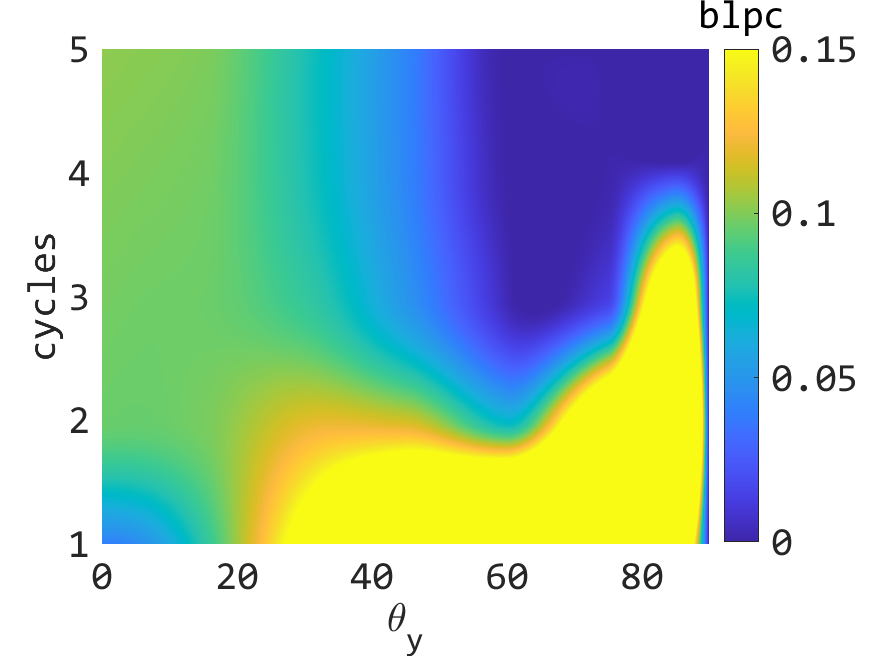}
         \caption{Field-induced MSRS: $\theta_\text{y}$}
         \label{fig:stabilityfieldy}
     \end{subfigure}
     \begin{subfigure}[b]{0.33\textwidth}
         \centering
         \includegraphics[scale=0.3]{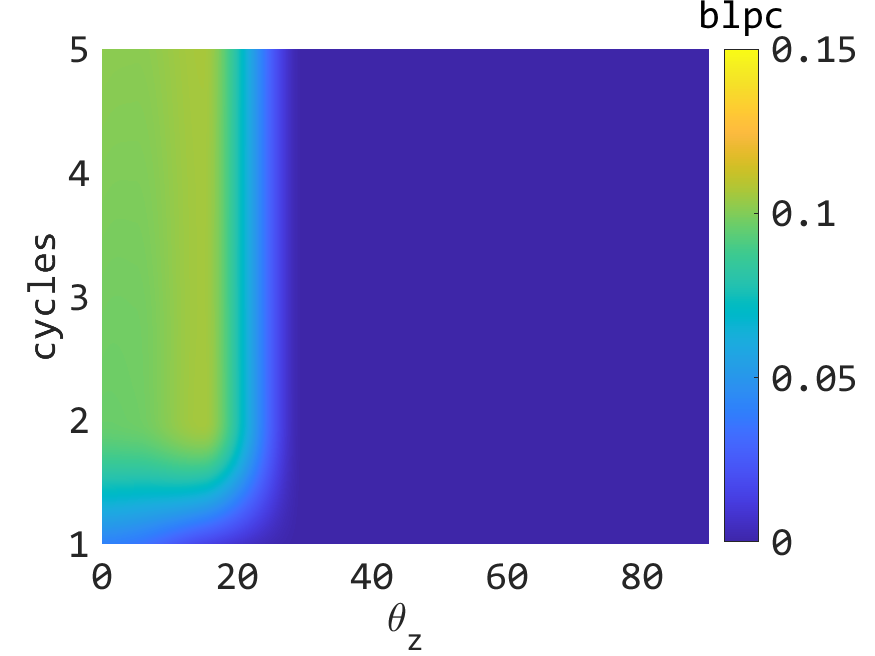}
         \caption{Field-induced MSRS: $\theta_\text{z}$}
         \label{fig:stabilityfieldz}
     \end{subfigure}
     \begin{subfigure}[b]{0.33\textwidth}
         \centering
         \includegraphics[scale=0.3]{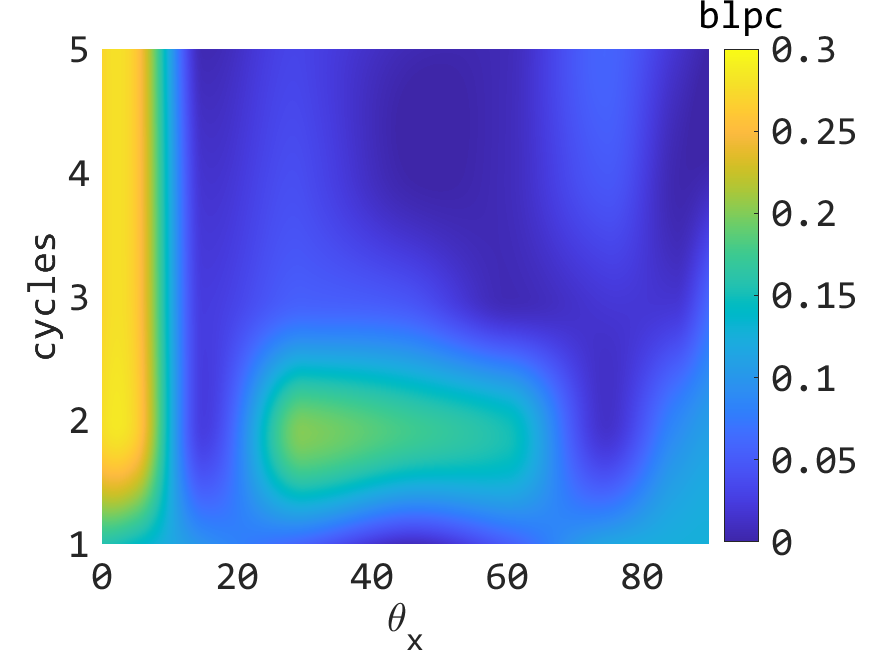}
         \caption{Finger-shaped MSRS: $\theta_\text{x}$}
         \label{fig:stabilityfingerx}
     \end{subfigure}
     \begin{subfigure}[b]{0.33\textwidth}
         \centering
         \includegraphics[scale=0.3]{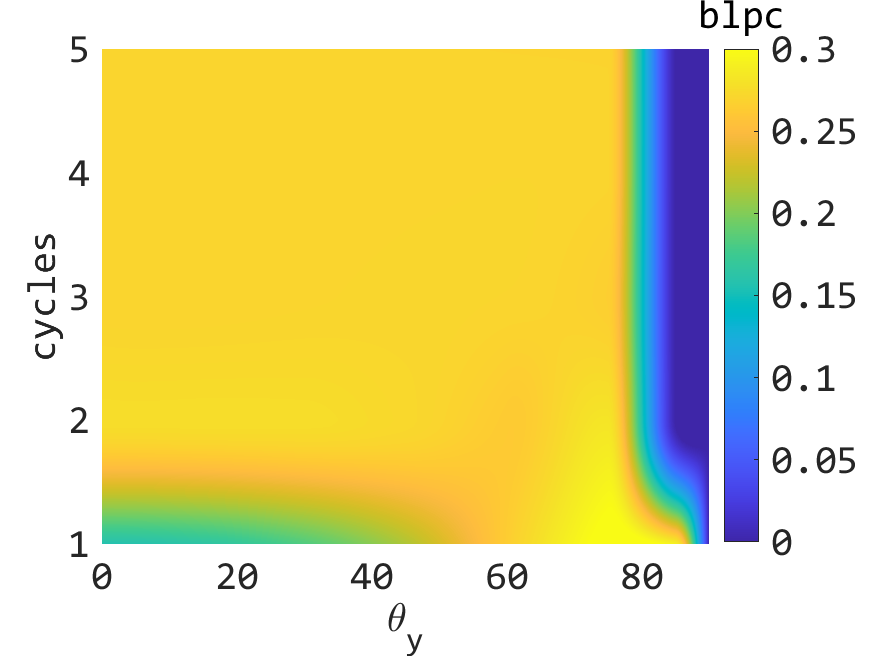}
         \caption{Finger-shaped MSRS: $\theta_\text{y}$}
         \label{fig:stabilityfingery}
     \end{subfigure}
     \begin{subfigure}[b]{0.33\textwidth}
         \centering
         \includegraphics[scale=0.3]{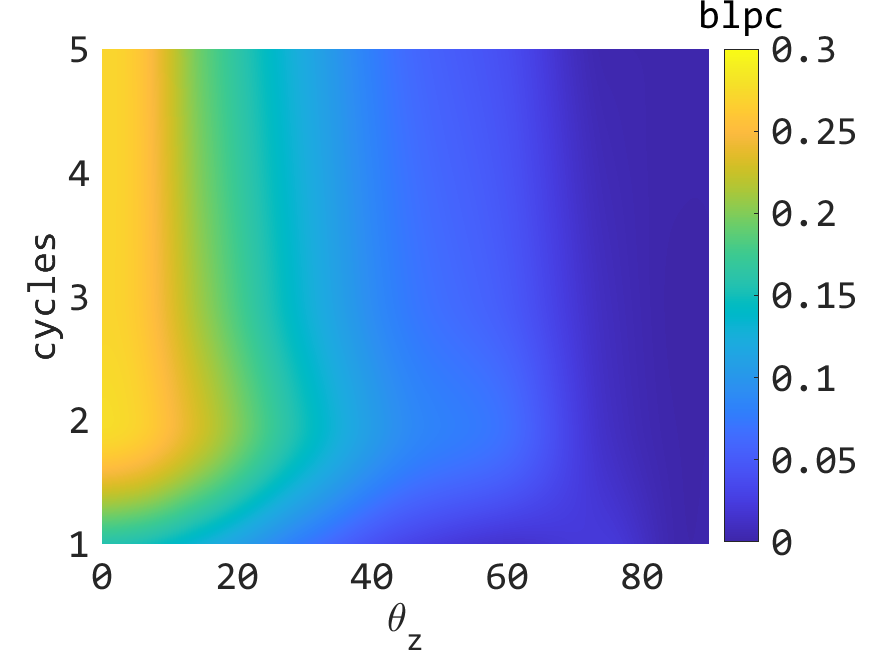}
         \caption{Finger-shaped MSRS: $\theta_\text{z}$}
         \label{fig:stabilityfingerz}
     \end{subfigure}
        \caption{Color maps have been used to represent the stability analysis for each swimmer considering three different starting configurations: roll (x-axis), pitch (y-axis), and yaw for full variation from 0 to 90 degrees.}
        \label{fig:stability}
\end{figure*}

Often during real-life applications, dynamic alterations occur due to changing flow fields. The swimmers often deviate from their perfect (initial) alignments as proposed earlier (see Fig. \ref{fig:swimmers}). This is when it becomes mandatory to investigate whether these swimmers are robust (and adaptive) enough to accommodate any orientational error (e.g., initial tilt) over the next few cycles to re-adjust themselves for further swimming. Therefore, we check this criterion by applying an initial tilt in their initial configuration without changing any other system parameters. Essentially, these swimmers are still subjected to the original magnetic fields. The initial tilt is classified into three categories for individual swimmers: roll (x-axis), pitch (y-axis), and yaw (z-axis). We plot the evolution of blpc over the first five swimming cycles for different tilt angles ranging from 0 to 90 degrees (individually for roll, pitch, and yaw) to see whether it gradually steadies into the steady-state swimming speed, or deviates from its desired motion (and then undergoes uncharacteristic swimming motion). Although the swimming responds differently to different initial tilts, we can still estimate the stability of each swimmer based on their blpc for the first five cycles (see Fig. \ref{fig:stability}).

We observe that the carangiform-like MSRS has the most uniform color palette, which means that the gradual transition to the steady-state (appropriate) orientation is achieved over the first few cycles irrespective of the initial tilt. We suggest referring to Supplementary Information for an animation (for better visualization) of adjustments over 5 cycles for a carangiform-like MSRS with initial spatial tilt. However, for all the other swimmers, either of the raw, pitch, or yaw (initial) tilts has some kind of non-uniformity in the swimmer blpc with an increasing number of cycles. For example, the initial 'pitch' tilt introduces non-uniform gradients in the color palette of the anguilliform-like MSRS in Fig. \ref{fig:stabilityanguilliformy}. Similar observations can be made for Figs. \ref{fig:stabilitydragz} ('yaw' tilt for drag-induced helical MSRS), \ref{fig:stabilityfieldy} ('pitch' tilt for field-induced helical MSRS), and \ref{fig:stabilityfingerx} ('roll' tilt for finger-shaped helical MSRS). These specific plots reveal that the swimmers become unstable (uncharacteristic swimming motion) either due to self-contact, coiling, floppy, or wobbly regimes. Only for the exception of the carangiform-like MSRS, we observe that it sustained the initial tilt for roll, pitch, and yaw, and could recover from its deviation of initial alignment over the first few swimming cycles. Consequently, we envisage this swimmer as the best candidate considering all the aspects of forward swimming, on-the-fly bi-directional locomotion, ease of maneuverability, and stability. For bi-directional swimming and steering in carangiform-like MSRS, refer to Supplementary Information for the respective animations.

\section{Conclusion}

In this study, we design magnetic soft robotic swimmers (abbreviated as MSRSs) for potential biomedical and microfluidic applications. We fabricate these active elastica and study their swimming kinematics under remote magnetic actuation. Silicone elastomers are impregnated with magnetic filler particles to render them magneto-responsive. Therefore, these smart soft composites deform and shape-morph when subjected to external magnetic fields in a pre-defined manner. When operated inside a viscous fluid medium, these miniaturized functional systems are observed to propel through the fluid using a combination of magnetic and viscous forces. Here, we thoroughly investigate helical and undulatory magnetically-actuated miniature soft robotic swimmers for efficient propulsion at low Re flows.

\begin{table*}[htbp]
\centering
\begin{tabular}{llllll}
\hline
\textbf{MSRS} & \textbf{Swimming mode} & \textbf{Max. blpc (M$_\text{n}$, F$_\text{n}$)} & \textbf{Bi-directionality} & \textbf{Stability analysis} \\
\hline\hline
Finger-shaped & Helical & 0.31 (191, 5) & No & Partially \\
\hline
Field-induced & Helical & 0.11 (398, 5) & Yes & Partially \\
\hline
Drag-shaped & Helical & 0.13 (53, 15) & No & Partially \\
\hline
Carangiform-like & Undulatory & 0.12 (526, 5) & Yes & Fully \\
\hline
Anguilliform-like & Undulatory & 0.12 (255, 5) & Yes & Partially \\
\hline
\end{tabular}
\vspace{0.3cm}
\caption{A comparison of different swimming modalities (i.e., the maximum value of blpc, on-the-fly bi-directionality, and stability) amongst all the MSRS.}
\label{table:comparison}
\end{table*}

We compare the kinematic performance (normalized steady-state swimming speed) of these magnetic swimmers with variations in non-dimensional magnetic and fluid numbers using an experimentally-calibrated computational framework. Our computational model simultaneously accounts for the magnetics, fluid dynamics, solid mechanics, and large deformation fluid-solid interaction to study the magnetic-field-induced propulsion and swimming kinematics of the proposed MSRSs. We analyze the influence of different aspect ratios and magnetic lengths on swimming behavior and kinematic performance. The optimal choices of remnant magnetization profiles and magnetic field actuation patterns for the highest swimming speeds for individual swimmers are reported. Furthermore, the maneuverability of these swimmers is investigated in terms of on-the-fly bi-directionality. Considering their real-life applications, we also consider the effect of initial tilt (i.e., different starting configurations) on the steady-state swimming behavior (and adjustment) with an increasing number of swimming cycles.

Finally, we compare all the MSRS based on different swimming modalities to identify the optimal swimmer (see Table \ref{table:comparison}). We note that the finger-shaped swimmer exhibits the highest blpc of 0.31; however, it cannot perform on-the-fly bi-directional swimming reversal. Furthermore, it showed a high steering radius, which meant that it was slow to take turns. As for the other helical swimmers, only the field-induced swimmer could swim in the reverse direction on-the-fly (subjected to opposite magnetic rotation directions). The drag-induced swimmer propelled along the same direction even when the field was reversed (due to the inversion of chirality upon inversion of the magnetic field rotation direction). Both these swimmers were also mostly stable for different initial starting orientations. We also observed that the undulatory swimmers were more stable compared to the helical swimmers. There was not much difference between the two undulatory swimmers in terms of blpc. Importantly, the carangiform swimmer proved to be the most versatile, since bi-directional swimming, and maximum stability were achieved for this MSRS. Therefore, this study points to the helical finger-shaped swimmer exhibiting the largest swimming speed, but the most versatile was observed to be the  carangiform-like MSRS (that had a simple oar-like motion mimicking midge larvae) that demonstrated the best swimming modalities.

\section{Materials and Methods}
\textit{Materials and Chemicals:} The following raw chemical agents have been used for the preparation of MSRSs: NdFeB magnetic filler particles (MQP-15-7, Magnequench), Ecoflex0010 (Smooth-On), tape (Magic tape, 3M), acrylic plate, Sweet Corn Syrup (CJ), and deionized water.

\textit{Preparation of MSRSs:} First, the NdFeB magnetic filler particles are mixed with Ecoflex0010 with a 1:1 weight ratio. This is followed by a vacuum deforming in a desiccator for 10 minutes. Then, five layers of tape are applied on the two edges and an uncured polymer mixture is poured between the tapes. Next, the mixture is gently sliced with a sharp blade to achieve a uniform thickness. The acrylic plate is then placed on the hot plate (60$^{\circ}$C, 1 hour) until the mixture is fully cured. The cured material is then laser cut (ProtoLaser U3, LPKF) to achieve the thin-uniform magnetic polymer sheet. Specifically, for the finger-shaped MSRS, the central protrusion (middle finger) is removed and replaced with pure Ecoflex before finally detaching the magnetic soft composite.

\textit{Characterization and Testing:} We develop a highly viscous fluid to ensure a low Reynolds number flow environment. Therefore, we mixed Sweet Corn Syrup with deionized water in a 20:1 volume ratio. Here, it is important to note that the magnetic phase distribution is spatially uniform to impart a net remnant magnetization equal to zero. Therefore, we finally magnetize the swimmer with a VSM machine (EZ7, Microsense) (1.2T, 6 seconds) to induce a magnetization profile. This entire fabrication procedure is schematically shown in Fig. \ref{fig:fabrication}. Before we expose the MSRSs to external magnetic coils for experiments, we conduct a calibration process for the coil system to minimize the magnetic field gradients and ensure the uniformity (spatial symmetry) of the magnetic field (Gaussmeter model 460, LakeShore). Later, we programmed the control signal and applied the desired magnetic fields using LabVIEW (NI 9269, National Instruments).

\textit{Solid-fluid Computational Fluid Dynamics approach:} We use a fully coupled computational model (based on an in-house finite element method) that simultaneously incorporates magneto-dynamics as well as large deformation FSI to study the complex interplay amongst elastic, viscous, and magnetic forces. Using an experimentally-calibrated computational framework that is based on a fictitious domain method (and solved in a monolithic manner) simultaneously accounting for large deformation FSI, magnetics, solid mechanics, and fluid dynamics, we study, compare, and optimize the swimming kinematics of MSRSs \cite{khaderi2012fluid}.

We first solve for the magnetic fields locally around the swimmer. These magnetic fields create body torques that deform the elastica, which in turn pushes the surrounding fluid to generate the thrust required for net propulsion. In the computational domain: the fluid phase is represented by an Eulerian mesh, while the solid by a Lagrangian mesh. The fluid-structure interaction (FSI) coupling has been done using Lagrange multipliers to ensure no-slip boundary conditions between the swimmer and the fluid. The swimmers are modeled using finite shell elements accounting for large deflections using an updated Lagrangian framework. Owing to miniaturized length scales (and therefore, low Reynolds numbers), we model the flow field using the Stokes equation, the solution of which is written using Green's functions in an infinite fluid domain \cite{blake1971note}. The drag forces on the swimmer are treated as a distribution of surface point forces. The fluid tractions (i.e., drag forces) are imposed upon the solid mechanics model of the swimmer as external body forces. The reader is referred to \cite{namdeo2014numerical,zhang2022metachronal,zhang2021transport,dong2020bioinspired,milana2020metachronal} for more details.

Briefly, default values of time step dt=0.1ms and mesh size ds=0.23mm were used throughout for all the swimmers considered. If these values are normalized with respect to the characteristic time (T=0.2s) and length (L=5mm) scales, then the number of time steps $n_t$ = T/dt = 2e3 and number of finite (constant strain) triangular elements $n_s$ = L/ds = 21.74; for details on the benchmark tests, convergence study, source code, and the underlying mathematical framework, the reader is referred to an authors' previous work \cite{khaderi2012fluid}.

\section*{Supporting Information}
A description of Supplementary Information and its contents (movies and animations) are available as attachments.

\section*{Acknowledgements}
The authors would like to thank the Center for Information Technology of the University of Groningen for their support and for providing access to the Habrok high-performance computing cluster. The work is carried out within the research program of the Centre for Data Science and Systems Complexity (DSSC), Faculty of Science and Engineering, University of Groningen.

\section*{Conflict of Interest}
The authors declare no conflict of interest.

\section*{Data Availability Statement}
The data that support the findings of this study are available from the corresponding author upon reasonable request.

\bibliographystyle{ieeetr}
\bibliography{sample}

\end{document}